\documentclass[a4paper, 11pt]{article}

\usepackage{ragged2e}
\usepackage{etoolbox}

\usepackage{xpatch}

\makeatletter

\xpatchcmd{\@makeschapterhead}{%
  \Huge \bfseries  #1\par\nobreak%
}{%
  \Huge \bfseries\centering #1\par\nobreak%
}{\typeout{Patched makeschapterhead}}{\typeout{patching of @makeschapterhead failed}}

\xpatchcmd{\@makechapterhead}{%
  \huge\bfseries \@chapapp\space \thechapter
}{%
  \huge\bfseries\centering \@chapapp\space \thechapter
}{\typeout{Patched @makechapterhead}}{\typeout{Patching of @makechapterhead failed}}

\makeatother

\usepackage{abstract}

\usepackage{placeins}

\usepackage[top=3cm,bottom=3cm,left=3cm,right=3cm,marginparwidth=2.5cm,headsep=11pt,headheight=14pt,a4paper]{geometry}

\usepackage[utf8]{inputenc}

\usepackage[dvipsnames]{xcolor}
\usepackage{hyperref}
\usepackage{mathtools}
\usepackage{graphicx}
\usepackage{booktabs}
\usepackage{stmaryrd}
\usepackage{color}
\usepackage{multirow}

\usepackage{amsmath,amsfonts,bm}

\newcommand{\G}{\mathcal{G}}

\renewcommand{\a}{\alpha}

\def\Id{{\bm{I}}}

\def\eqref#1{equation~\ref{#1}}

\def\1{\bm{1}}

\def\vh{{\bm{h}}}

\def\vs{{\bm{s}}}

\def\vu{{\bm{u}}}

\def\vx{{\bm{x}}}

\def\mC{{\bm{C}}}

\def\mJ{{\bm{J}}}

\def\mU{{\bm{U}}}

\DeclareMathAlphabet{\mathsfit}{\encodingdefault}{\sfdefault}{m}{sl}
\SetMathAlphabet{\mathsfit}{bold}{\encodingdefault}{\sfdefault}{bx}{n}

\DeclareMathOperator{\Tr}{Tr}

\renewcommand{\a}{\alpha}

\newcommand{\trs}{^{\dagger}}

\usepackage[]{appendix}
\usepackage{caption}
\usepackage{subcaption}
\usepackage{natbib}
\usepackage{bibentry}
\newcounter{bibcount}
\makeatletter
\patchcmd{\@lbibitem}{\item[}{\item[\hfil\stepcounter{bibcount}{\thebibcount.}}{}{}
\renewcommand\NAT@bibsetup%
  [1]{\setlength{\leftmargin}{\bibhang}\setlength{\itemindent}{-\parindent}%
      \setlength{\itemsep}{\bibsep}\setlength{\parsep}{\z@}}
\makeatother

\usepackage{rotating}
\usepackage{bbm}
\usepackage{latexsym}
\usepackage{footmisc}
\usepackage{authblk}

\usepackage{cases}

\begin{document}

\title{Optimal regularizations for data generation with probabilistic graphical models}

\author{A. Fanthomme, F. Rizzato, S. Cocco, R. Monasson}
\affil{Laboratory of Physics of the Ecole Normale Sup\'erieure, \\ PSL Research and CNRS UMR8023, 24 rue Lhomond, 75005 Paris, France }

\maketitle

\begin{abstract}
Understanding the role of regularization is a central question in Statistical Inference. Empirically, well-chosen regularization schemes often dramatically improve the quality of the inferred models by avoiding overfitting of the training data. We consider here the particular case of $L_2$ and $L_1$ regularizations in the Maximum A Posteriori (MAP) inference of generative pairwise graphical models. Based on analytical calculations on Gaussian multivariate distributions and numerical experiments on Gaussian and Potts models we study the likelihoods of the training, test, and `generated data' (with the inferred models) sets as functions of the regularization strengths. We show in particular that, at its maximum, the test likelihood and the `generated' likelihood, which quantifies the quality of the generated samples, have remarkably close values. The optimal value for the regularization strength is found to be approximately equal to the inverse sum of the squared couplings incoming on sites on the underlying network of interactions. Our results seem largely independent of the structure of the true underlying interactions that generated the data, of the regularization scheme considered, and are valid when small fluctuations of the posterior distribution around the MAP estimator are taken into account. Connections with empirical works on protein models learned from homologous sequences are discussed.
\end{abstract}


\section{Introduction}

Data-driven modeling is now routinely used to address hard challenges in an increasing number of fields of science and engineering for which first-principle approaches have limited success. Applications include the characterization and design of complex materials \citep{schmidt_recent_2019}, shaped by the pattern of strong and heterogeneous interactions between their microscopic components.  Performance of data-driven models strongly depends on the choice of their hyperparameters, such as the architecture, and the strengths of the regularization penalties. These parameters are generally set through empirical procedures, such as cross-validation with respect to a goodness-of-fit estimator. Unfortunately, this common approach often offers no insight about why these values of the parameters are optimal, and may not guarantee that the obtained models are well-behaved with respect to other estimators. This paper reports some efforts to address these issues for the specific case of $L_p$-norm regularization and probabilistic graphical models.

Probabilistic graphical models rely on the inference of the set of conditional dependencies between the variables under study, which, in turn, may be used to generate new configurations of these variables \citep{mackay_information_2003}. Regularization allows the graph of pairwise conditional dependence to satisfy some properties of interests, such as to be sparse or to have dependence factors bounded from above. Among the huge variety of applications of those models, substantial efforts have been devoted over the past decades to applications to the modeling of proteins based on homologous, i.e. evolutionary related sequence data. Unveiling the relations between the functional or structural properties of a protein and the sequence of its amino acids is a difficult task. Graphical model-based modeling consists of inferring a graph of effective interactions between the amino acids, which reproduce the low-order (1- and 2-point) statistics in the sequence data; for reviews, see \citep{cocco_inverse_2018} for protein modelling and \citep{berg_17} for general inference of graphical models with discrete variables. In practice, for proteins with few hundreds of amino acids, tens of millions of interaction parameters have to be inferred. To avoid overfitting, regularization of those interactions, often based on pseudocounts, or $L_1$- and $L_2$--norms are generally introduced, with intensities varying with the optimality criteria chosen by the authors \citep{barton_14, Haldane_2019}. For instance, Ekeberg et al. chose regularization strength scaling linearly with the number of data (sequences) \citep{ekeberg_improved_2013, ekeberg_fast_2014} to maximize the quality of structural predictions. Hopf et al. chose  linear scaling with the  dimension of the data (sequence length) and with the number of possible amino-acid types (generally, $q=20$) for predicting the fitness effects resulting from mutations along the sequence \citep{hopf_mutation_2017}. The rationale for these scalings and what they tell us about the underlying properties of the protein system remains unclear. In addition, whether these scalings are appropriate for generating new data points, i.e. for the design of new protein sequences having putative properties is not known, and other regularization schemes have been proposed \citep{barrat2021}

In the following, we propose to study the role of regularization in the inference process, replacing Potts models by Multivariate Gaussian models in order to make the problem analytically tractable in some limiting cases. We show that two natural definitions for the optimal values of the regularization strength are in practice very close to one another, and that their common value can be related to the amplitude of the ground-truth interactions, in agreement with experimental observations. Our paper is organized as follows. In Section 2, we introduce the Gaussian model and the regularizations of interest. Numerical results are reported in Section 3. Section 4 is devoted to the analytical studies of the poor and excellent sampling limits. Last of all, some conclusions and perspectives are drawn in Section 5.

\section{Gaussian Vectors Model and Regularization}

\subsection{Expression of likelihood in the large--size limit}

In order to be able to model distributions over $n$--dimensional vectors, we consider first the multidimensional Gaussian distribution, often referred to as Gaussian Vectors or Spherical Model. In the following, we will only consider the case of centered Gaussian Vectors, for which the mean value of each component vanishes and the probability density is given by:
\begin{equation}
    p(\vx)= \frac{1}{\sqrt{(2 \pi)^n \det(\mC^{tr})}} e^{-\frac{1}{2} \vx^T (\mC ^{tr})^{-1} \vx} \ , \label{gaussian_map:eq:multivariate_gaussian_pdf}
\end{equation}
where $\mC^{tr}$ is the $n\times n$--dimensional covariance matrix.
Alternatively we may define the underlying data distribution through an interaction matrix $\mJ^{tr}$, which represents the interaction strength between the variables (vector components). This interaction matrix $\mJ^{tr}$ is related to the true covariance matrix $\mC^{tr}$ of the data through
\begin{equation}
    \mC^{tr} = (\mu^{tr} \Id - \mJ^{tr})^{-1}\ ,
\end{equation}
where $\mu^{tr}$ was introduced to impose the spherical normalization constraint $Tr(\mC^{tr})=n$.
Denoting as $(j^{tr}_1, \dots, j^{tr}_n)$ the eigenvalues of $\mJ^{tr}$, the normalization condition can be written, in the large $n$ limit, as
\begin{equation}
    1 - \frac{1}{n}\sum_{k=1}^n \frac{1}{\mu^{tr}-j^{tr}_k} = 0. \label{gaussian_map:eq:normalization}
\end{equation}
As the covariance matrix is non-negative we are looking for the unique value of $\mu^{tr}$ in $[\max_k \{ j^{tr}_k\}, +\infty[$ that satisfies this equation.

In the following, we will be interested in inferring the  interaction matrix $\mJ^{tr}$ from an empirical approximation $\mC^{emp}$ of the correlation matrix obtained using $p=\a \, n$ samples $(\vx^1, \dots, \vx^p)$ as:

\begin{equation}
    \forall (i,j) \in [1, n]^2, \, \mC^{emp}_{i,j} = \frac{1}{p} \sum_{k=1}^p \vx^k_i \vx^k_j\ .
\end{equation}
We define the posterior likelihood of any interaction matrix $\mJ$ given the empirical covariance matrix $\mC^{emp}$,
\begin{equation}\label{eq:Bayes}
    p(\mJ|\mC^{emp}) = e^{-n\,E(\mJ)}\ ,
\end{equation}
where the energy function $E(\mJ)$ reads
\begin{equation}
    E(\mJ) =  - \frac{\alpha}{2} Tr(\mJ \mC^{emp}) + \alpha \log Z(\mJ) + \frac{\gamma}{4} Tr(\mJ^2). \label{gaussian_map:eq:energy}
\end{equation}
In the expression above the first two terms correspond to the standard likelihood of a given Gaussian Model given the empirical covariance, while the last term expresses a penalty on the $L_2$ norm of the inferred interaction matrix. The strength of this regularization is controlled by the parameter $\gamma$.

The partition function $Z(\mJ)$ of the so-called spherical spin model reads
\begin{equation}
\begin{split}
    \log Z(\mJ) &= \int_{\vx\in \mathbb{R}^n} \delta( \vx^2=n) \, e^{\frac{1}{2}\sum_{i\neq j} x_i J_{ij} x_j} \\
&= n \min_{\mu} \left ( \frac{\mu}{2} - \frac{1}{2n} \log(\det(\mu\Id-\mJ)) \right ) \label{eq:logZ}
\end{split}
\end{equation}
to the dominant order in $n$.
The parameter $\mu$ can be interpreted as a Lagrange multiplier, introduced to impose the spherical constraint $Tr(\mC)=n$, which corresponds exactly to the normalization condition~(\ref{gaussian_map:eq:normalization}) but with the eigenvalues of the true interaction matrix $j^{tr}$ replaced by the ones of $\mJ$.

Our goal will be to minimize the energy~(\ref{gaussian_map:eq:energy}) with respect to the interaction matrix $\mJ$; the matrix $\mJ^*$ minimizing the energy will be called inferred matrix and will be our primary object of study. We also define $\mu^*$ the Lagrange multiplier imposing the spherical constraint on this inferred model, and $\mC^*$ the covariance matrix of the inferred model. For reference, we define in Table~\ref{gaussian_map:tab:all_quantities} all the different quantities that we will be considering and their associated notations.

\begin{table}
    \centering
    \begin{tabular}{c c}
        Symbol & Quantity \\
        \toprule
          $\Id$ & The identity matrix \\
          $n$ & Dimension of the Gaussian Vectors \\
          $p$ & Number of samples \\
          $\a$ & Sampling ratio $p/n$ \\
          $\gamma$ & The strength of the $L_2$ penalty \\
          $\mJ$ & Dummy variable standing for an interaction matrix \\
          $\mC$ & Dummy variable standing for a covariance matrix \\
          $\mJ^{tr}$ & True interaction matrix of the underlying model \\
          $\mC^{tr}$ & True covariance matrix of the underlying model \\
          $\mC^{tr, rot}$ & True covariance matrix, in the diagonalizing basis of $\mC^{emp}$ \\
          $c^{tr}$ & An eigenvalue of the true covariance matrix \\
          $\mu^{tr}$ & Lagrange multiplier imposing the spherical constraint on $\mJ^{tr}$ \\
          $\mC^{emp}$ & Empirical covariance matrix obtained from $p=\alpha n$ samples\\
          $c^{emp}$ & Eigenvalue of the empirical covariance matrix \\
          $\mJ^*$ & Interaction matrix obtained from Maximum A Posteriori inference \\
          $j^*$ & Eigenvalue of the MAP inferred interaction matrix \\
          $\mu^*$ & Lagrange multiplier imposing the spherical constraint on $\mJ^*$ \\
          \end{tabular}
    \caption{All quantities used in the inference procedure. Please note that the empirical covariance matrix $\mC^{emp}$ and its eigenvalues are stochastic quantities for a given underlying interaction model $\mJ^{tr}$ (since they depend on the exact samples drawn). Additionally, we will assume the eigenvalues $c$ to be ordered from largest to smallest, and denote with a lower--index $k$ both $c^{emp}_k$ (the $k$--th largest eigenvalue of $\mC^{emp}$) and $j^*_k$ the corresponding eigenvalue of $\mJ^*$ (see eqn.~\ref{gaussian_map:eq:MAP_equation_eigenvalues}). }
    \label{gaussian_map:tab:all_quantities}
\end{table}

\subsection{Maximum A Posteriori estimator of the interaction matrix}

When $\gamma$ is equal to $0$, the regularization disappears and the Maximum Likelihood estimation of $\mJ^*$ is exactly equal to the one computed from the empirical covariance $\mC^{emp}$; when $\gamma$ goes to infinity, the regularization becomes so strong that the inferred interaction matrix is exactly equal to $\bm{0}$; in the general case of finite $\gamma$, we find $\mJ^*$ by computing $\frac{\partial E}{\partial \mJ}(\mJ^*)$, which yields the Maximum A Posteriori (MAP) equation:

\begin{equation}
\gamma \mJ^* - \a \mC^{emp} + \a (\mu^* \Id-\mJ^*)^{-1}=0.
    \label{gaussian_map:eq:MAP_equation}
\end{equation}

According to equation~(\ref{gaussian_map:eq:MAP_equation}) the inferred interaction matrix $\mJ^*$ is diagonal in the same vector basis as the empirical covariance matrix $\mC^{emp}$. It is therefore possible to rewrite this equation in terms of the eigenvalues (respectively, $j^*$, $c^{emp}$) of those matrices\footnote{Because of equation~\ref{gaussian_map:eq:MAP_equation}, we know that to each eigenvalue of the empirical covariance matrix corresponds exactly one eigenvalue of the inferred interaction matrix.}:
\begin{equation}
    \gamma \, {j^*}^2 - (\gamma\mu^* +\alpha c^{emp})\, j^* + \alpha (\mu^* c^{emp} -1) = 0.
\end{equation}

Since the discriminant $\Delta=(\alpha c^{emp}-\gamma \mu^*)^2 + 4 \a\gamma \geq 0$, the eigenvalue $j^*(c^{emp})$ always exists in $\mathbb{R}$ and is found to be equal to:
\begin{equation}
    j^*(c^{emp}) = \frac{1}{2\gamma}\left (\a c^{emp} + \gamma \mu^* - \sqrt{(\a c^{emp} - \gamma\mu^*)^2+4\a\gamma}\right).
\label{gaussian_map:eq:MAP_equation_eigenvalues}
\end{equation}

It should be noted here that this is in fact an auto-consistent equation: $\mu^*$ is used to compute the eigenvalues $j^*$, which in turn are used to compute $\mu^*$. In order to solve it, we consider $\mu^*$ to be a free parameter and make the expression of the inferred eigenvalues depend on two variables $j^*(c^{emp}, \mu^*)$. Introducing the corresponding expression into the normalization condition~\ref{gaussian_map:eq:normalization}, we find that $\mu^*$ is the only root\footnote{It can easily be shown that $\partial j^*(c^{emp}) / \partial \mu^*$ is always positive; since $j^*(c^{emp}, \mu) < \mu$, we have that $Res(\mu)$ is well-defined for all values of $\mu$; $\partial Res(\mu) / \partial \mu$ is always positive and therefore $Res(\mu)$ is monotonically increasing from $-\infty$ when $\mu \rightarrow -\infty$ to $1$ when $\mu \rightarrow +\infty$, ensuring the unicity of the root.} of the residual function:

\begin{equation}
    Res^{norm}(\mu) = 1 - \frac{1}{n} \sum_k \frac{1}{\mu - \frac{1}{2\gamma}\left (\a c^{emp}_k + \gamma \mu - \sqrt{(\a c^{emp}_k - \gamma\mu)^2+4\a\gamma}\right)}.
\end{equation}

In practice, the optimization of this residual is performed numerically in Python using the Van Wijngaarden-Dekker-Brent method \cite{brent_algorithms_2013}, implemented within the SciPy package \cite{virtanen_scipy_2020}. After obtaining the value of $\mu^*$, the inferred interaction matrix $\mJ^*$ is obtained by computing its spectrum through equation~\ref{gaussian_map:eq:MAP_equation_eigenvalues} and changing the basis back from the inference basis (which diagonalizes the empirical covariance $\mC^{emp}$) to the original basis (in which the true interaction $\mJ^*$ was defined).

\subsection{Likelihoods of the training, test, and generated sets}

In order to be able to compare the quality of the inferred interaction matrix $\mJ^*$ as a function of the different parameters of the system (namely, $\a$, $\gamma$ and the true interaction matrix $\mJ^{tr}$) the first interesting quantity to define is the training likelihood:
\begin{equation}
    L_{train} = \frac{1}{p}\sum_{k=1}^p \left [ \frac{1}{2} \sum_{i, j} J_{ij}^* \, x^k_i x^k_j - \log Z(\mJ^*) \right ],
\end{equation}
which directly quantifies how well the examples of the training set are fit by the MAP estimator $\mJ^*$. By performing the summation over the sample index $k$, the likelihood can be rewritten as a function of the empirical covariance matrix $\mC^{emp}$:
\begin{equation}\label{eq:ltrain}
    L_{train} = \frac{1}{2} \sum_{i, j} J_{ij}^* C^{emp}_{ij} - \log Z(\mJ^*).
\end{equation}

A similar reasoning can be performed, this time considering the case where an infinite number of samples are drawn from the true underlying distribution (meaning that $C^{emp}$ is replaced by $\mC^{tr}$), corresponding to the average test error on samples independent of the training ones. This leads to the definition of the test likelihood:
\begin{equation}\label{eq:ltest}
    L_{test} =\frac{1}{2} \sum_{i, j} J_{ij}^* C^{tr}_{ij} - \log Z(\mJ^*).
\end{equation}

 Finally, one can also consider the likelihoods of a `generated set' of examples drawn using the inferred interaction matrix, with respect to this same inferred interaction matrix $\mJ^*$:

 \begin{equation}
      L_{gen} =\frac{1}{2} \sum_{i, j} J_{ij}^* C^{*}_{ij} - \log Z(\mJ^*).
 \end{equation}
It is possible to rewrite the "generated" likelihood using the MAP equation:
\begin{equation}
\begin{split}
    L_{gen} &=\frac{1}{2} \sum_{i,j} J_{ij}^* C^{*}_{ij} - \log Z(\mJ^*)
            =\frac{1}{2}\sum_{i,j} J_{ij}^* \frac{1}{\mu\Id-\mJ^*}\Big|_{ij} - \log Z(\mJ^*)\\
            &\stackrel{(\ref{gaussian_map:eq:MAP_equation})}{=}\frac{1}{2}\sum_{i,j} J_{ij}^* (\mC^{emp}_{ij}-\frac{\gamma}{\a}\mJ^*_{ij}) - \log Z(\mJ^*)
            = L_{train} - \frac{\gamma}{2\a} \sum_{i,j} {J_{ij}^*}^2. \\
\end{split}
\label{gaussian_map:eq:gen_train_equality}
\end{equation}

\noindent This form of the generated likelihood can be interpreted as a form of bias-variance trade-off: if an increase in the magnitude of the couplings is necessary to better fit the training set, it will increase the variance of the generated data and consequently decrease the generated set likelihood.

\subsection{Generic dependence of the likelihoods upon regularization strength}

\begin{figure}
    \centering
    \includegraphics[width=\textwidth]{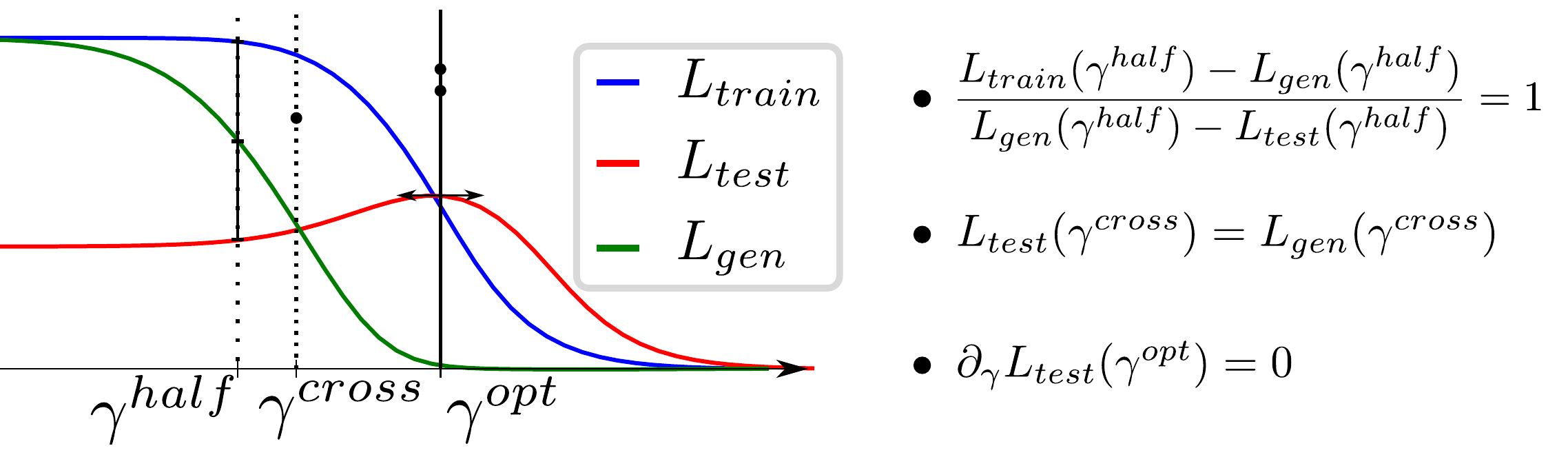}
    \caption[Definition of the noticeable gammas]{Sketch of the expected behaviours of the likelihoods vs. regularization $\gamma$, and definitions of the three values of interest: $\gamma^{half}$, for which the generated likelihood is exactly in-between the train and test ones; $\gamma^{cross}$, for which the test and generated likelihoods are equal; $\gamma^{opt}$, for which the test likelihood is maximal. The difference between optimal and crossing likelihoods is strongly exaggerated for illustration purposes, as in practice they are found to be extremely close to each other in almost all circumstances.}
    \label{gaussian_map:fig:noticeable_gammas}
\end{figure}

Figure~\ref{gaussian_map:fig:noticeable_gammas} is a sketch of the typical behaviours expected for the three log-likelihoods defined above as the  regularization strength $\gamma$ is varied:

\begin{itemize}
    \item For weak regularization {\em i.e.} $\gamma$ close to zero MAP inference is unconstrained, and the inferred covariance coincides with the empirical one. The value of the training likelihood is large, as the details of the training set are fitted. Consequently, the inferred model has poor generalization capability, and the test log-likelihood has a low value. This is a situation of \textit{overfitting}. Generated data look like training data, so the generated likelihood is large.
    \item For strong regularization, {\em i.e.} large $\gamma$ the regularization term in the energy becomes more important than the likelihood term, so that the MAP estimator $\mJ^*$ tends to zero; this is a case of \textit{under fitting}, as the training, test, and generated likelihoods will be low. When $\gamma$ goes to infinity, the three likelihoods converge to a common value,
    \begin{equation}\label{eq:common}
        L{(\gamma\to \infty)} = - \frac{n}{2}.
    \end{equation}

\item In-between those two regimes, {\em i.e.} for intermediate values of $\gamma$ the training likelihood is monotonically decreasing with $\gamma$, reflecting the increasing bias towards small couplings, and so is the generated likelihood. The test likelihood displays a non-monotonic evolution, and reaches a maximum for some regularization penalty $\gamma^{opt}$. While the presence of $\gamma$ biases the inference, it also reduces its variance, and hence allows for better generalization of the model to unseen examples. While the test likelihood always remains smaller than the training likelihood (as should be expected, the model cannot generalize better than it fits the available data), the test and generated likelihoods may cross at a certain value $\gamma^{cross}$,  Figure~\ref{gaussian_map:fig:noticeable_gammas}. We also define  the regularization  $\gamma^{half}$ for which the generated likelihood is half way between the train and test ones.
\end{itemize}
In the following we will study, through numerical experiments and analytical calculations the behaviour of these three regularization strengths of interest, and their dependence on the model defining parameters (number $p$ of samples compared to the size $n$, structure of the coupling matrix, ...).

\section{Numerical experiments}

\subsection{Gaussian Vectors Model}

In order to study the dependence of $\gamma^{opt}, \gamma^{cross}, \gamma^{half}$ with the different parameters, we implemented the MAP inference procedure in Python (the code is available on  \href{https://github.com/AFanthomme/Gaussian-Model-Likelihoods}{GitHub}).

The general procedure is as follows: first, an interaction matrix $\mJ^{tr}$ is randomly generated, according to an underlying distribution (see next subsections for details on the distributions we considered); then, a certain number $p=\alpha \, n$ samples are drawn from the Gaussian Vectors model with interactions $\mJ^{tr}$, and from those samples an empirical covariance matrix $\mC^{emp}$ is derived; this matrix is then diagonalized, and the spectrum of the MAP interaction estimator $\mJ^*$ is computed through eqn.~(\ref{gaussian_map:eq:MAP_equation_eigenvalues}); the training and generated set likelihoods are computed directly using those eigenvalues, while the test likelihood requires the inversion of the diagonalization basis change in order to obtain the expression of $\mJ^*$ in the same basis as $\mC^{tr}$.\footnote{Those two basis \textit{a priori} coincide if and only if $\a \rightarrow \infty$.}

\subsubsection{Case of random quenched couplings}\label{sec:rqc}

\paragraph{The condensation phase transition.} We assume that the underlying interaction matrix is drawn from the Gaussian Orthogonal Ensemble, \textit{i.e.} all its components are drawn at random and independently from a centered Gaussian distribution:
\begin{equation}
    \forall \, i, j \, , \quad J^{tr}_{ij} \sim \G\big(0, \frac{\sigma}{\sqrt{n}}\big). \label{gaussian_map:eq:gaussian_matrix}
\end{equation}
The presence of this $1/\sqrt{n}$ normalization ensures that the energy is extensive with $n$. The model is "infinite range" because all spins are interacting with all other spins with similar strengths, controlled by the parameter $\sigma$. As shown in \cite{kosterlitz_spherical_1976} the model exhibits a condensation phase transition when $\sigma$ crosses the critical value $\sigma_c=1$. For $\sigma>\sigma_c$ one eigenvalue of the  covariance matrix scales linearly with $n$, while all others remain finite. This transition can be intuitively  understood as follows. Since the  interaction matrix $\mJ^{tr}$ has Gaussian entries, its eigenvalue distribution follows Wigner's semi-circle law, and ranges from $-2\sigma$ and $2\sigma$. As $\sigma$ increases from small values, the value of the Lagrange multiplier $\mu$ imposing the spherical constraint becomes closer and closer to its lower-bound $2\sigma$, and the gaps closes (in the infinite $n$ limit) when $\sigma=\sigma_c$. For $\sigma>\sigma_c$ $\mu$ remains equal to $2\sigma$, and the corresponding top eigenvector of $\mJ^{tr}$  gives rise to an extensively large eigenvalue in $\mC^{tr}$. More precisely, when $\sigma$ is larger than $\sigma_c$, the maximum eigenvalue of $\mC^{tr}$ is equal to
\begin{equation}
   c^{tr}_{max} =n \times\left(1- \frac{1}{2\pi\sigma^2} \int^{2\sigma}_{-2\sigma} \frac{\sqrt{4\sigma^2-j^2}}{2\sigma-j} dj \right) = n\;\left(1-\frac 1\sigma\right).\label{eq:max_eig_ferro}
\end{equation}
In this situation, the model  generates configurations that are effectively constrained close to a subspace of dimension $1$.

\paragraph{Evolution of the log-likelihoods with $\gamma$.}

\begin{figure}
    \centering
    \includegraphics[width=\textwidth]{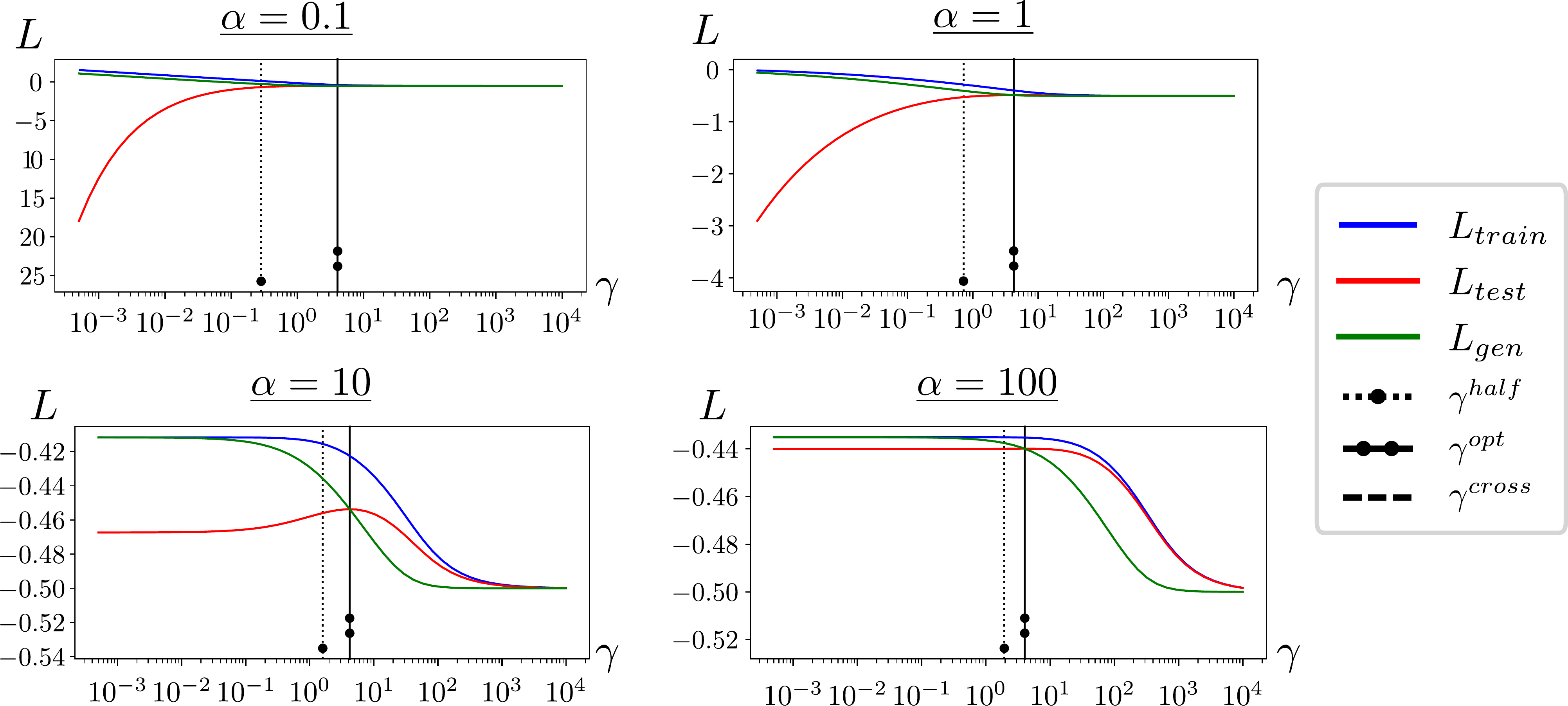}
    \caption[Typical evolution of the likelihoods]{Evolution of the four likelihoods (normalized by $n$) as functions of the regularization strength $\gamma$ for four different values of the sampling ratio $\a$. In all cases, both training and generated likelihoods are monotonically decreasing, while the test likelihood is first increasing then decreasing; the training and test likelihoods never cross, while the generated and test likelihoods cross for a value of the regularization extremely close to the optimum of $L_{test}$.}
    \label{gaussian_map:fig:likelihoods_behavior}
\end{figure}

Figure~\ref{gaussian_map:fig:likelihoods_behavior} shows the behaviours of the log-likelihoods with varying $\gamma$, for different regimes of low and high sampling fractions $\alpha$. Vertical lines locate the three values of $\gamma$ of interest. The overall shape of the curves agree with the expected behaviours sketched in Figure~\ref{gaussian_map:fig:noticeable_gammas}.

For small $\gamma$ (overfitting regime), the value of the training likelihood is very large, irrespective of the value of $\a$ as the weak regularization allows the inference procedure to fit the training set without bias. The test loss, however,  strongly varies $\a$. For low sampling (small $\a$) $\mC^{emp}$ is essentially uncorrelated with $\mC^{tr}$, and the test likelihood will be very low. If $\a$ is large, $\mC^{emp}$ is almost equal to $\mC^{tr}$, and the test likelihood will be very close to its training counterpart, both being very high. In all cases the generated and the training log-likelihoods coincides.

When $\gamma$ is very large, the regularization term in the energy pushes the MAP estimator $\mJ^*$ towards 0. In this \textit{underfitting} regime, all log-likelihoods tend to the same limit value, see eqn.~(\ref{eq:common}).

\begin{figure}
    \centering
    \includegraphics[width=\textwidth]{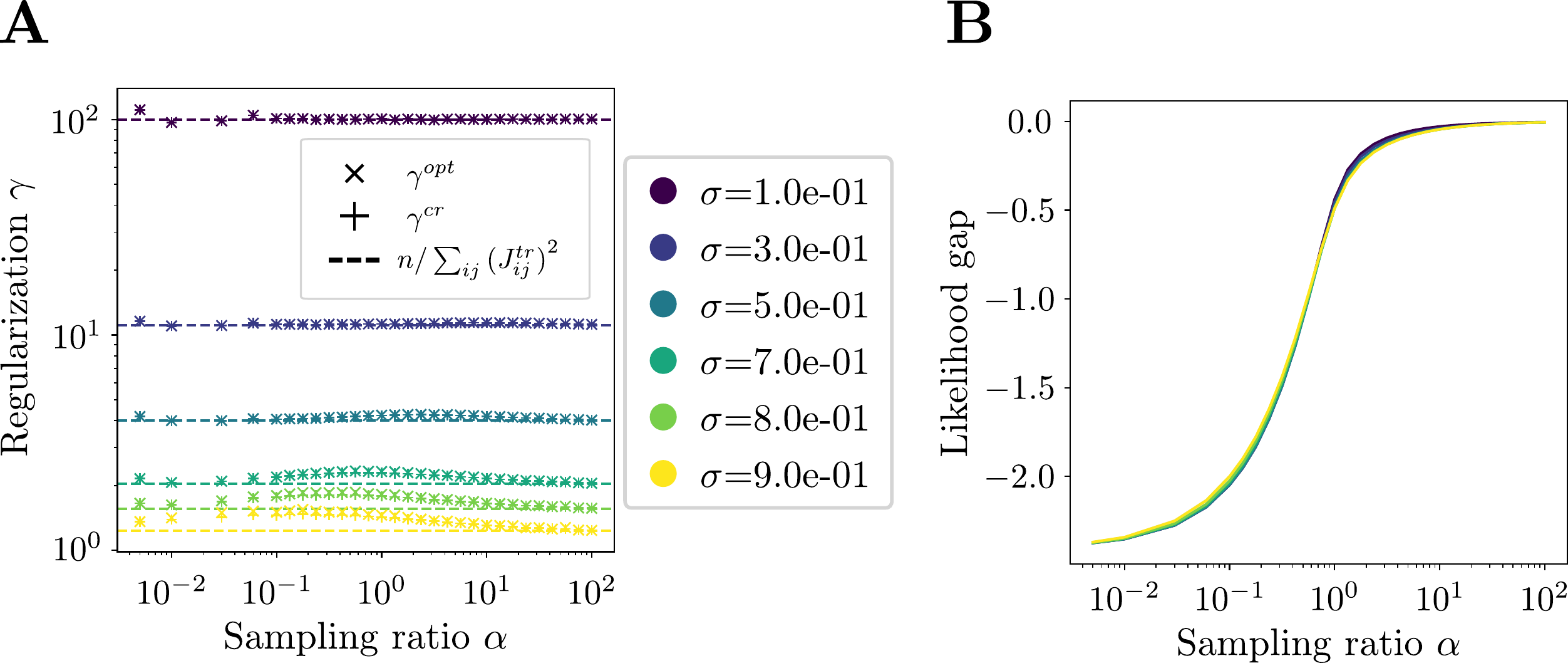}
    \caption[Evolution of the likelihoods]{Gaussian Vectors Model with $L_2$ regularization. \textbf{A}: Evolution of the regularizations $\gamma^{opt}$ and $\gamma^{cross}$ as functions of the sampling ratio $\a$ for different values of the interaction dispersion $\sigma$, see eqn.~(\ref{gaussian_map:eq:gaussian_matrix}). The theoretical prediction for $\gamma^{cross}$, represented here as a dashed line for each value of $\sigma$, is given in eqn.~(\ref{eq:pred}) and derived in Section~\ref{sec:analytical}. \textbf{B}: Evolution of the likelihood gap $\Delta L = L_{test}(\a, \gamma^{opt}(\a)) - L_{train}(\a=\infty, \gamma=0)$ as a function of $\a$ for the same values of $\sigma$ as panel \textbf{A}. As expected, this gap vanishes as $\a$ goes to infinity, meaning that the optimal inferred model (obtained with non-zero regularization) fits the data perfectly in the limit of infinite samples. While the gaps are identical between different values of the interaction strength, we were not able to determine the expression for this evolution.}
    \label{gaussian_map:fig:likelihood_evolution}
\end{figure}

For intermediate $\gamma$, we observe that the location of the maximum of the test likelihood, $\gamma^{opt}$, is very close to the value of the regularization strength $\gamma^{cross}$ for which it crosses the generated log-likelihood. This unexpected results holds in most circumstances as reported in Figure~\ref{gaussian_map:fig:likelihood_evolution}, but small discrepancies can be observed  at low sampling ratio $\alpha$. Detailed analytical calculations  for the Gaussian Vectors Model in Section~\ref{sec:analytical} will allow us to confirm this numerical observation, and, in addition, to approximate their common value as a function of the sampling ratio $\alpha$ and of the "true" interaction matrix $\mJ^{tr}$:
\begin{equation}\label{eq:pred}
     \gamma^{opt}\simeq\gamma^{cross}\simeq \frac {n}{\sum_{i,j} (J_{ij}^{tr})^2}\ .
\end{equation}

Let us notice that $\gamma^{half}$, the regularization penalty at which the generated likelihood is the mean of the train and test ones, seems to approximately fullfill the following equality
\begin{equation}
    \sum_{i,j} J_{ij}^*(\gamma^{half})\, C^{tr}_{ij}= \sum_{i,j} J_{ij}^{tr}\, C^{*}_{ij}(\gamma^{half}) \ . \label{gaussian_map:eq:gamma_half_likelihoods_equality}
\end{equation}

In order to estimate $\gamma^{opt}$, $\gamma^{cross}$ and $\gamma^{half}$ as precisely as possible, we define in Appendix \ref{app:residuals} functions whose roots correspond to those regularizations, and optimize them numerically with care.

\subsubsection{Other types of underlying interactions}
\begin{figure}
    \centering
    \includegraphics[width=\textwidth]{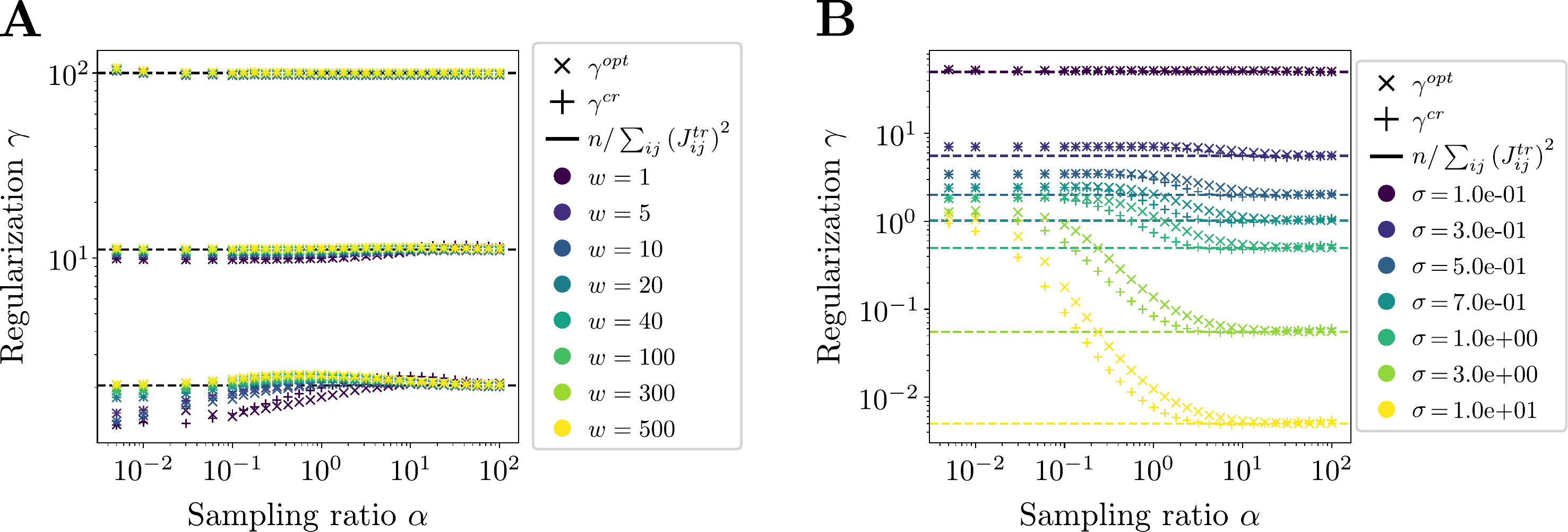}
    \caption[Influence of structured interactions]{Evolution of the  regularizations of interest for two different cases of structured interaction matrices $\mJ^{tr}$. \textbf{A}: case of a random band matrix. \textbf{B}: case of a deterministic, uniform $1$-dimensional chain. In both cases, the observation that the crossing and optimal regularizations are of the same order of magnitude remains true, and so does the prediction for their value in the $\a \rightarrow \infty$ regime.}
    \label{gaussian_map:fig:bands_tridiag}
\end{figure}

The empirical coincidence between $\gamma^{opt}$ and $\gamma^{cross}$ reported above extends to other choices of the coupling matrices. As an illustration we consider the case where the underlying interaction matrix $\mJ^{tr}$ is structured, instead of being randomly drawn. In particular, we present in Figure~\ref{gaussian_map:fig:bands_tridiag} two examples, and show that the presence of structure does not significantly alter our previous observations:
\begin{itemize}
    \item in panel \textbf{A}, the interaction matrix is band-diagonal, meaning that the coefficients are given by
    \begin{equation}
        \forall \, (i,j)\ s.t.\ |(i-j)\mod n|<\frac w2 \, , \quad J^{tr}_{ij} \sim \G\left(0, \frac{\sigma}{\sqrt{w}}\right) \ ,
    \end{equation}
  where $w$ is the width of the non-zero band, $\G$ is the Gaussian distribution,  and $[n]$ represents the 'modulo n' operation. This means that sites are arranged on a ring, with interactions only between $w$ nearest neighbors, and the value of those non-zero interactions are drawn randomly from a Gaussian distribution.

    This model can be related to the random Schrödinger operator in dimension 1, an object extensively studied in the context of Anderson localization, see \cite{anderson_absence_1958}. As observed numerically by \cite{casati_scaling_1990} and later rigorously proved (see \cite{bourgade_random_2018} for an overview), a phase transition can be observed when $w \sim \sqrt{n}$ between a regime (small $w$) where the eigenvectors of $\mJ^{tr}$ are localized \textit{i.e.} decay exponentially with distance, and another where they are extended (large $w$).

    Our particular choice of scaling of the individual entries of those band matrices is such that $\frac {\sum_{i,j} (J_{ij}^{tr})^2}{n}$ remains constant, and so do the expected values of the regularizations of interest.

    \item in panel $\textbf{B}$, $\mJ^{tr}$ is a deterministic matrix corresponding to a unidimensional chain:
     \begin{equation}
         \forall \, (i, j) \, , \quad J^{tr}_{ij} = \begin{dcases}
             &0 \text{ if $i=j$ or $|(i-j) \ \text{mod} \ n|>1$} \\
             &\sigma \text{ if $|(i-j)\ \text{mod}\ n|=1$ }
         \end{dcases},
     \end{equation}
    meaning that sites are again arranged on a ring, this time with fixed positive interactions between direct neighbors only. This particularly simple model does not  exhibit any phase transition.

\end{itemize}

We find that changing the underlying model of interaction does not significantly impact the phenomenology that we previously observed for infinite-range Gaussian interactions: an optimal regularization still exists for all values of the sampling ratio $\alpha$.

\subsubsection{$L_1$ regularization}

While the $L_2$ penalty is often used in practice, and encourages smoothness of the energy landscape, it is not the only possible choice. In many cases, it can be interesting to infer sparse interactions models, which is usually done by using an $L_1$ regularization: in a protein, amino acids which are very distant in the sequence can end up close in the folded structure, and therefore interact strongly so that one has to \textit{a priori} allow interactions between all sites along the sequence; however, in three-dimensional space, each site is close only to a very small fractions, so that the inferred interaction matrix should be sparse.
The inference procedure in this case is less straight-forward than for the $L_2$ case, and analytical solutions cannot be obtained in the general case. Instead, one relies on the so-called "Graphical Lasso" method \cite{friedman_sparse_2008}, which iteratively solves Lasso problems for each column of the interaction matrix using coordinate descent \cite{wright_coordinate_2015} until convergence, implemented in Scikit-learn \cite{pedregosa_scikitlearn_2011}.

\begin{figure}
    \centering
    \includegraphics[width=\textwidth]{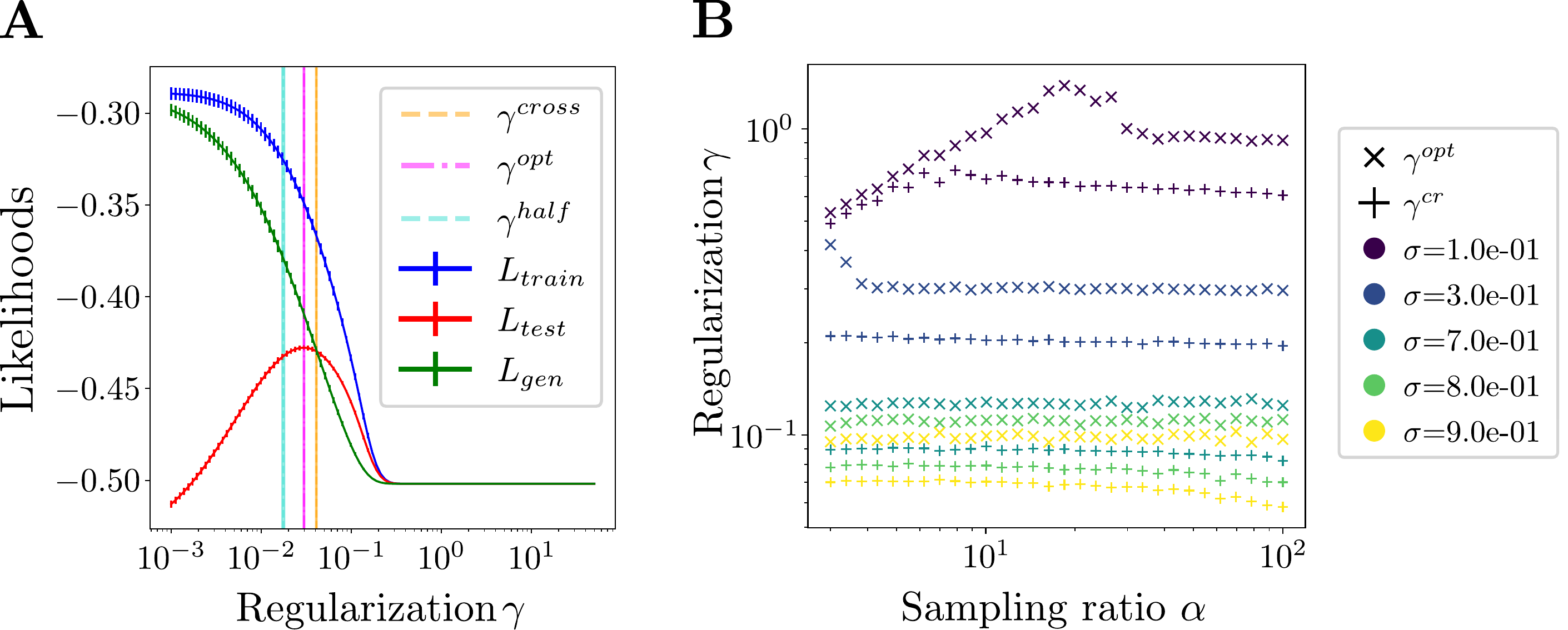}
    \caption[Extension to $L_1$ regularization]{\textbf{A}: Typical evolution of the likelihoods as a function of the strength of the $L_1$ regularization. The existence of a finite optimal regularization, as well as the crossing between test and generated likelihoods, remains true as in the $L_2$ case. \textbf{B}: Evolution of the crossing and optimal regularizations as a function of the sampling ratio $\alpha$. While the two noticeable regularizations are no longer equal, they remain of a similar order of magnitude. }
    \label{gaussian_map:fig:l1_summary}
\end{figure}

We show in Figure~\ref{gaussian_map:fig:l1_summary} that the behavior of the likelihoods remains qualitatively similar to what we observed in the case of $L_2$ regularization, despite the difference between the two noticeable regularizations being much higher than previously; similarly, the equality (\ref{gaussian_map:eq:gamma_half_likelihoods_equality}) that was observed at $\gamma^{half}$ remains close to being true, albeit less closely followed than in the $L_2$ case. A detailed analysis of this inference procedure could both shed light on the difference between the two, and give us a theoretical prediction for the optimal regularization in this regime, but this remains to be done in future work.

\subsection{Potts Model}

\subsubsection{Generation of synthetic data and energy model}\label{sec:ER}

We now consider a discrete-valued graphical model, in which each (categorical) variables may take one out of $q$ values. The energy of a configuration $\vx$ is given by
\begin{equation}\label{eq:potts1}
    E(\vx;\vh,\mJ) = -\sum_{i<j} J_{ij} (x_i,x_j) -\sum_i h_i(x_i)\ ,
\end{equation}
The local fields $\vh$ and the couplings $\mJ$ are, respectively, $q$--dimensional vectors and $(q\times q)$--dimensional matrices. The corresponding partition function is
\begin{equation}
    Z(\vh,\mJ) = \sum _{\{x_i=1,2,...,q\}} e^{-E(\vx;\vh,\mJ)}
\end{equation}

We start by drawing the components of $\vh^{tr}$ and $\mJ^{tr}$ that from Gaussian distributions of zero mean and standard deviations $\sigma_h^2$ and $\sigma_J^2$. All components of the $h$ vectors and $J$ matrices are chosen at random and independently from each other.

Next, each element of the Gaussian matrix $J^{tr}_{ij}$ is multiplied by a connectivity indicator equal to 0 or 1, which identifies, respectively, the absence or the presence of an edge between the variables $i$ and $j$ in the coupling network. In practice, we choose this connectivity at random,  following the prescription of the so-called  Erd\"os-R\'enyi (ER) random graph ensemble. For each pair $i,j$ of variables we chose to insert an edge in the interaction graph  with probability $d/n$, and to have no connection with probability $1-d/n$; $d/n\times(n-1)$ is therefore the average degree of each variable in the connectivity graph.

In our simulations, we vary
\begin{itemize}
\item the size (number of variables), $n$; here $n=25,50,100,150$;
\item the number of Potts states, $q$ (here $q=10,20$);
\item the probability $d/n$ to include  edges in the ER graph. Different values of $d$ were tested only for $n=25$, for which the computation were faster: $d=1.25,2.5,7,10$.
\end{itemize}

For each system, a number $p$ of data point, ranging from $10^2$ to $10^5$ were generated by Markov Chain Monte Carlo sampling. Intuition about the sampling level can be obtained by comparing $p$ with the number of parameters to infer from the data, $n\times q + \frac 12 n(n-1) \times q^2$. The parameters defining the Gaussian distribution to generate fields and coupling are here kept constant as $n$ varies: $\sigma^2_h=5$, $\sigma^2_J=1$.


\subsubsection{Behaviours of the train, test, and generated log-likelihoods}

Once the data are generated through Monte Carlo sampling of the Gibbs distribution associated to the energy (\ref{eq:potts1}) we infer the model parameters $h_i(x), J_{ij}(x,x')$ using two methods. The first one is the Pseudo-Likelihood Method (PLM), a non-Bayesian inference method that bypass the (intractable) computation of the partition function $Z$ \citep{ravikumar_highdimensional_2010,ekeberg_improved_2013}. The second one is the so-called Adaptive Cluster Expansion (ACE) algorithm, which recursively computes better and better approximations for the cross-entropy of the data (and $\log Z$) \citep{cocco_adaptive_2011,barton16}, combined with color compression \citep{Rizzato2020}.

The inference is done with a $L_2$-norm regularization on the couplings (intensity $\gamma$) and on the fields (intensity $\gamma_h$). We expect regularization to be much less needed for the fields, because single-site frequencies are much better sampled than pairwise frequencies. We therefore fix the ratio between the regularization of fields and couplings, setting $\gamma_h= \gamma/(10\,n)$, and vary $\gamma$.

\begin{figure}
    \centering
    \includegraphics[width=.8\textwidth]{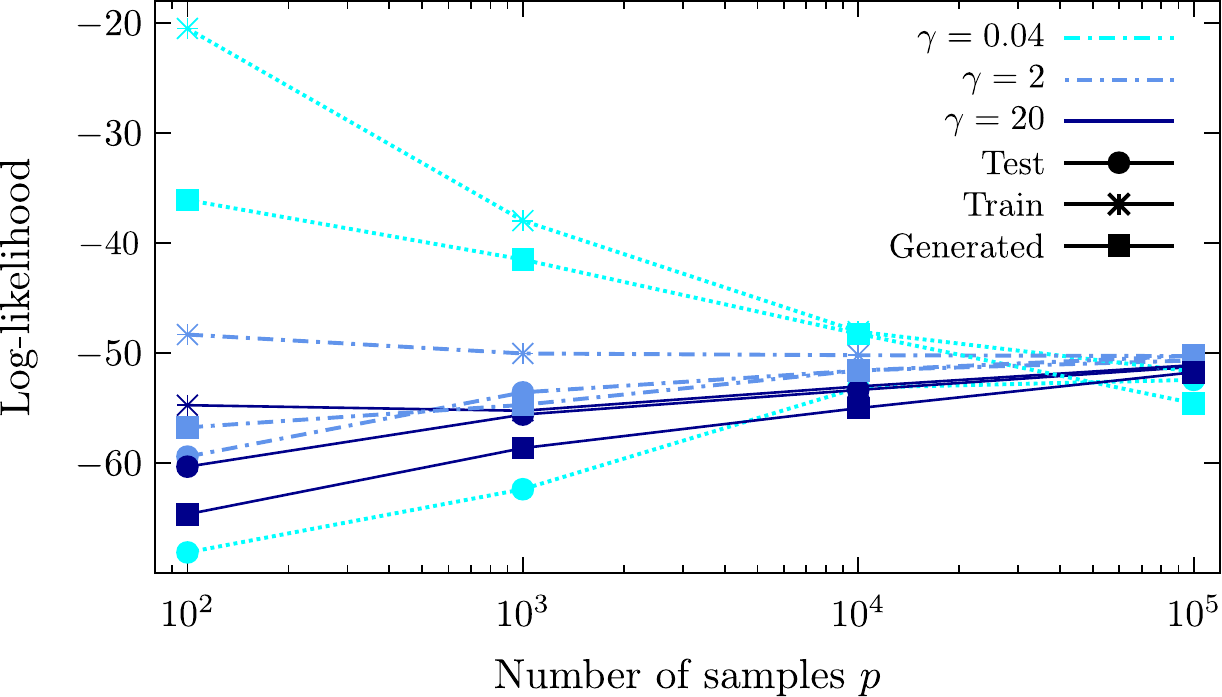}
	\caption{Average log-likelihoods of test (circular markers), train (cross markers) and generated (square markers) data vs. number $p$ of samples for different regularization strengths $\gamma$ (one color and line style for each). Results were averaged over 20,000 sequences for each reported value of $\gamma$ and $p$. Parameters: $n=50$, $q=10$.} \label{fig:PLM_logLinf}
\end{figure}

In Figure \ref{fig:PLM_logLinf}, we show the average log-likelihoods (normalized by $n$) of the data in the training set, in the test set (same size as the training set) and the generated data set. Model parameters were inferred with the PLM procedure, and log-likelihoods (and the log partition function) were computed with the Annealed Importance Sampling method. For small regularization $\gamma$ we observe a strong overfitting effect as expected, with similar values for $L_{train}$ and $L_{gen}$, much above $L_{test}$. For intermediate regularization values, the test and generated log-likelihoods are similar as the number $p$ of samples available for the inference increases, while the size $n$ is kept fixed. This result is compatible with a weak dependence of $\gamma^{cross}$  upon $\alpha$, as found  for the Gaussian Vectors Model. For large $\gamma$, $L_{gen}$ may get smaller than $L_{test}$, a signature of very strong underfitting.

\subsubsection{Dependence of optimal regularizations on system and data set sizes}
\label{sec:results}

We assess the quality of the inference through the Kullback-Leibler (KL) divergence of the inferred probability distribution from the ground-truth probability distribution,
\begin{equation}
    D_{KL} = \sum _{\vx} \frac{e^{-E(\vx;\vh^*,\mJ^*)}}{Z(\vh^*,\mJ^*)}\;\log\left[\frac{e^{-E(\vx;\vh^*,\mJ^*)}}{Z(\vh^*,\mJ^*)}\bigg/\frac{e^{-E(\vx;\vh^{tr},\mJ^{tr})}}{Z(\vh^{tr},\mJ^{tr})}\right]\ .
\end{equation}
Again, we estimate the partition functions entering the definition above with Annealed Importance Sampling.

\begin{figure}
\centering
	\includegraphics[width=\linewidth]{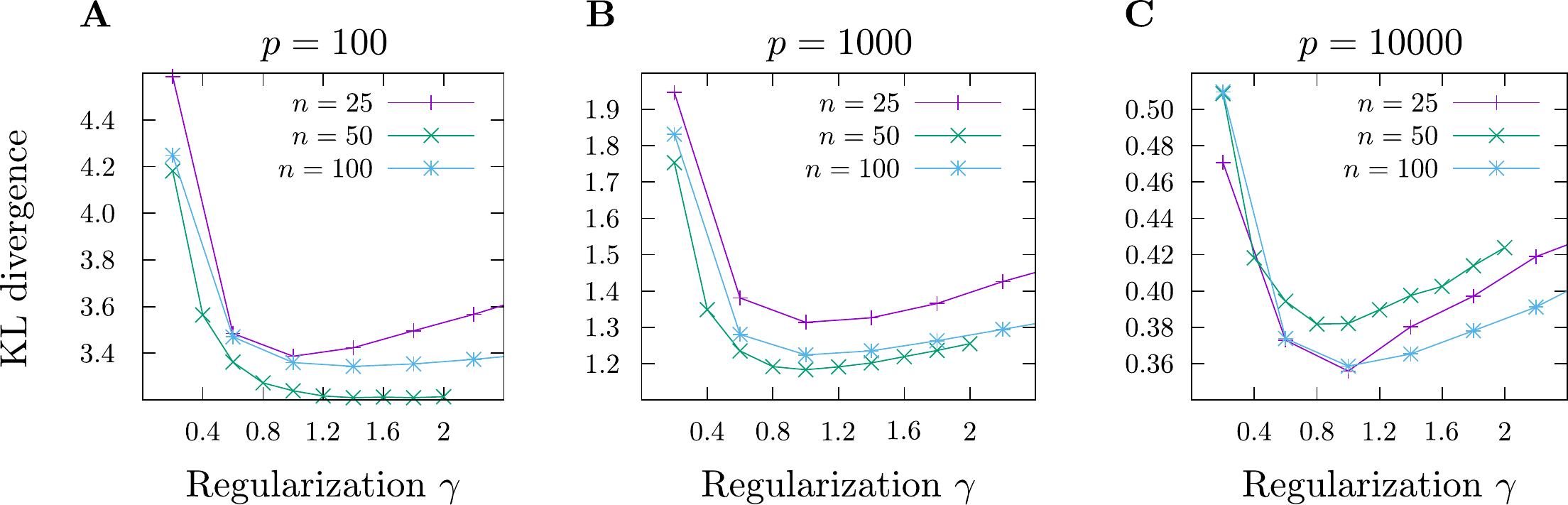}
 \caption{Kullback-Leibler (KL) divergence between the inferred models and the ground truth for different graph ($n$) and sampling ($p$) sizes as a function of the regularization on the couplings ($\gamma$). The $y$-axis was arbitrarily rescaled between the different curves to allow for easier comparison. Parameters: $d=2.5$, $q=10$. \label{fig:variousN}}
\end{figure}

\begin{figure}
\centering
	\includegraphics[width=\linewidth]{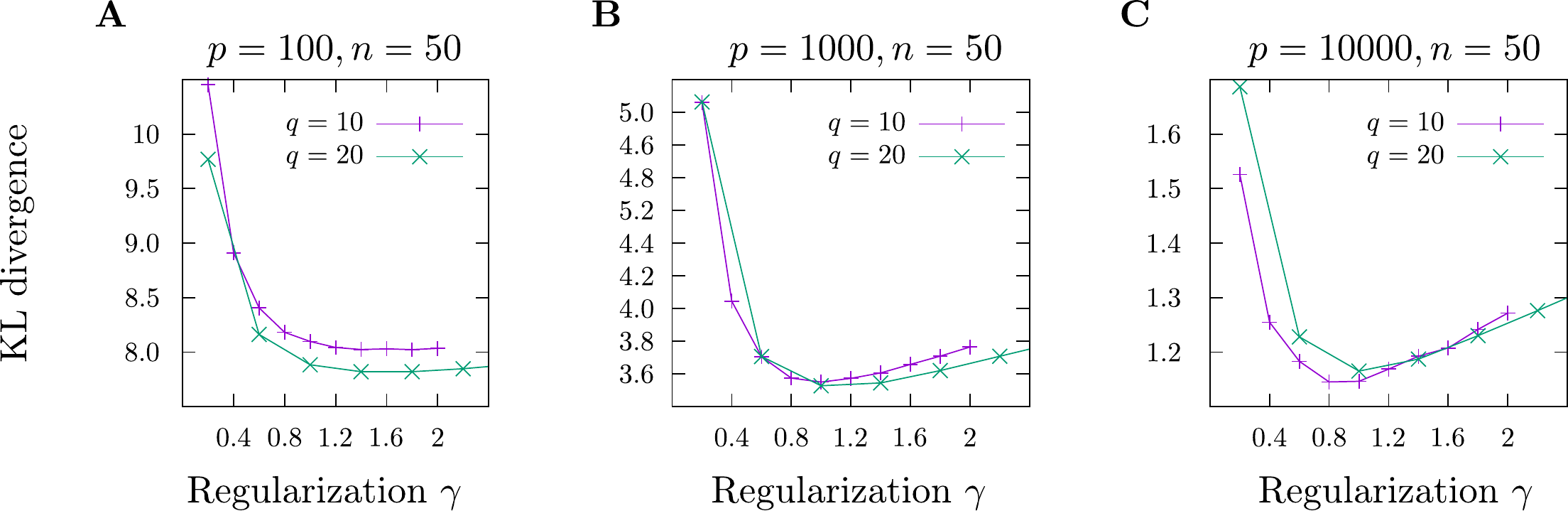}
 \caption{Kullback-Leibler (KL) divergence between the inferred models and the ground truth for different numbers $q$ of Potts states and $p$ of data points, as a function of the regularization on the couplings ($\gamma$) and for diff. The $y$-axis was arbitrarily rescaled between the different curves to allow for easier comparison. Parameters: $d=2.5$, $n=50$. \label{fig:variousQ}}
\end{figure}

\paragraph{Dependence on the size $n$.}\label{sec:constdensity}
We first study if and how the optimal regularization parameter $\gamma$ changes when we the system size $n$ is increased, while the average connectivity in the graph is fixed by choosing $p=2.5/n$;  we also fix the number of Potts states to $q=10$. In Figure~\ref{fig:variousN} we show the KL divergence for models inferred at different $\gamma$ for various $n$ and $p$. The optimal regularization $\gamma^{opt}$ seems to be roughly equal to $0.5$ in all the considered cases, independently of $p$ (with some inaccuracy for very poor sampling, {\em i.e.} $p=100$). We have also checked that this optimal value of $\gamma$ does not seem to depend on $q$, by repeating the same numerical experiments for $q=20$ Potts states with similar results, see Figure~~\ref{fig:variousQ}.

These two results are in very good agreement with the theoretical prediction reported in eqn.~(\ref{eq:pred}), that is, $\gamma^{opt}\simeq \gamma^{cross}\simeq \frac 1d = 0.4$ for the parameters chosen in Figures~\ref{fig:variousN} and \ref{fig:variousQ}. Indeed, in ER graphs, the average number of interacting neighbours is equal to $d$ (on average), independently of $n$ (and $p$). In addition, since each variable can take one out $q$ symbol values, the number of variables $j$ interacting with $i$ in the sum at the denominator in eqn.~(\ref{eq:pred}) is independent of $q$.

\begin{figure}
\centering
	\includegraphics[width=\linewidth]{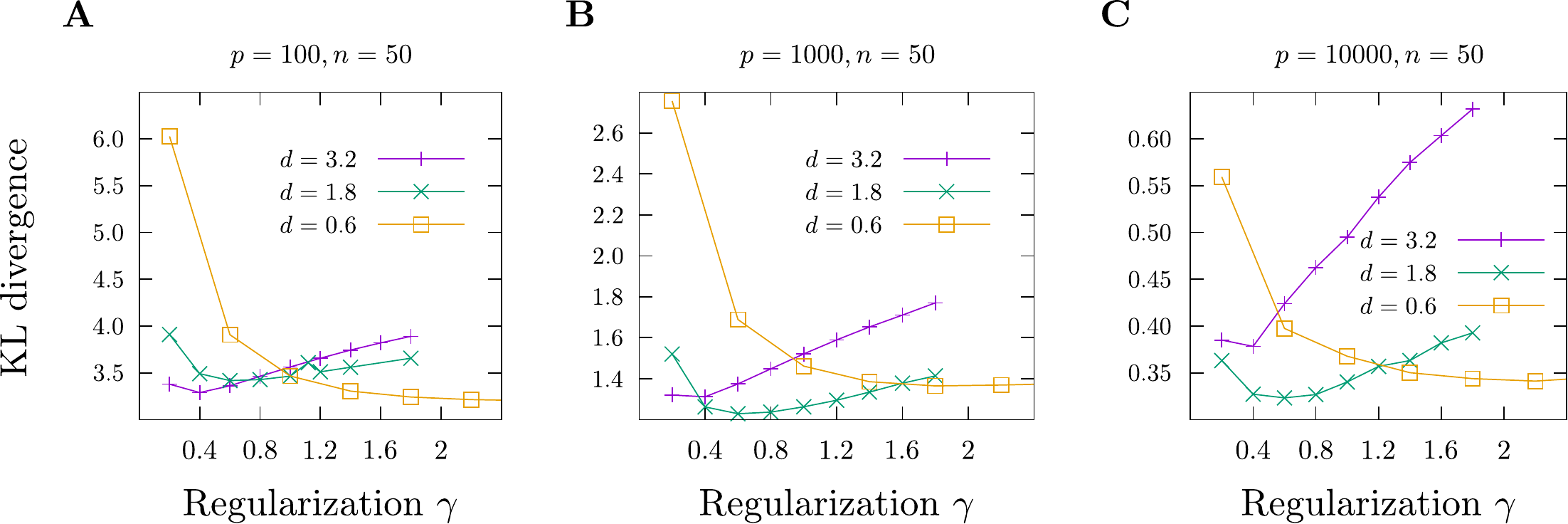}
	\caption{Kullback-Leibler (KL) divergence between the inferred models and the ground truth for different average number of edge per site (numbers reported in the panels, obtained by varying $d$), as a function of the regularization ($\gamma$) used during inference. The $y$-axis was arbitrarily rescaled between the different curves to allow for easier comparison. Parameters: $n=25$, $q=10$.
 	\label{fig:variousdensity}}
\end{figure}

\paragraph{Dependence on the structural connectivity of the interaction graph.}\label{sec:diffdensity}
We then study how the optimal regularization depends on the connectivity of the graph. For this reason we keep the graph size fixed ($n=25$), and build different ER models with different densities varying $d$, see section \ref{sec:ER}. Once data are generated we infer the model parameters $\vh^*,\mJ^*$ for different $\gamma$ and sample sizes $p$. Results are reported in Figure \ref{fig:variousdensity}, and show a clear dependence on the structural parameter $d$.  We observe that the scaling factor is approximately inversely proportional to the number of neighbors on the interacting graph. This result is in excellent agreement with the outcome of the expected theoretical scaling reported in eqn.~(\ref{eq:pred}).

\section{Analytical calculations at low and high sampling ratios}
\label{sec:analytical}

While finding the exact value for the regularization strengths of interest as functions of the model parameters is out of reach we show in this section how this calculation can be done in the case of the Gaussian Vectors Model for very low and high values of the sampling ratios.

\subsection[Asymptotic behavior of the crossing regularization]{Asymptotic behavior of $\gamma^{cross}$ }

\label{gaussian_map:sec:asymptotic_gamma_cross}

The crossing regularization $\gamma^{cross}$ is defined through
\begin{equation}
    L_{test}(\gamma^{cross}) = L_{gen}(\gamma^{cross})\ .\label{gaussian_likelihood:eq:gamma_cross_condition}
\end{equation}
Replacing $L_{gen}$ in the equation above with its expression in  eqn.~(\ref{gaussian_map:eq:gen_train_equality}) and using the definitions (\ref{eq:ltrain},\ref{eq:ltest}) of the train and test log--likelihoods we obtain
\begin{equation}\label{eq:gc2}
\gamma^{cross} = \alpha\;  \frac{L_{train}(\gamma^{cross}) -
    L_{test}(\gamma^{cross}) }{ \sum_{i,j} {J^*_{ij}}^2} = \alpha \; \frac{
     \sum_{i,j} J^*_{ij}\big(  C^{emp}_{ij}- C^{tr}_{ij} \big)}{ \sum_{i,j} {J^*_{ij}}^2}\ .
\end{equation}

\subsubsection{$\a \rightarrow \infty$ regime}

We derive below an asymptotic prediction for $\gamma^{cross}$ in the large sampling regime $\a \rightarrow \infty$. We begin by considering the $\alpha \gg 1$ limit of the matrix-form MAP eqn.~(\ref{gaussian_map:eq:MAP_equation}):
\begin{equation}
    \mJ^* = \mu \Id - \big({\mC^{emp}}\big)^{-1}. \label{gaussian_map:eq:MAP_equation_asymptotic_high_alpha}
\end{equation}
We consider the distribution of the empirical covariance matrix $\mC^{emp}$ conditioned to the "true" correlation matrix $\mC^{tr}=(\mu^{tr}-\mJ^{tr})^{-1}$, known as the Wishart distribution \cite{wishart_generalised_1928}, and defined for $p>n$ as
\begin{equation}
p_{\mJ^{tr}}(\mC) \propto e^{n \frac{\a}{2}\mathcal{F}(\mC)}\ , \quad  \mathcal{F}(\mC)= \frac{\a-1}{2} \log\det(\mC) - \frac{\a}{2}\Tr\Big((\mu^{tr}-\mJ^{tr})\mC \Big)
\end{equation}
where we omit $\mC$-independent normalization factor.
For large $\a$, we can perform a saddle-point approximation of this density around its maximum $\mC^{tr}$:
\begin{equation}\label{eq:lk1}
    p_{\mJ^{tr}}(\mC=\mC^{tr} + \Delta \mC) \propto e^{n \frac{\a}{2}  \Delta \mC\trs \frac{\partial^2 \mathcal{F}}{\partial \mC \partial \mC}(\mC^{tr}) \Delta \mC} .
\end{equation}
A straightforward calculation leads to
\begin{equation}
 \frac{\partial^2 \mathcal{F}}{\partial C_{i,j} \partial C_{a,b}}(C^{tr}) =  \frac{\partial^2 \log\det{\mC}}{\partial C_{i,j} \partial C_{a,b}}(\mC^{tr}) = - \big({C^{tr}}\big)^{-1}_{a,i} \, \big({C^{tr}}\big)^{-1}_{b,j} \ .
\end{equation}
We deduce from eqn.~(\ref{eq:lk1}) that $\big(\mC^{tr}\big)^{-1}\times\Delta \mC=\mU/\sqrt{n\alpha}$, where $\mU$ is distributed as an uncorrelated Gaussian matrix, whose entries have zero means and unit standard deviation. Therefore, using eqn.~(\ref{gaussian_map:eq:MAP_equation_asymptotic_high_alpha}), we have
\begin{equation}
    \mJ^* \simeq \mu\Id - \big( \mC^{tr} +\Delta \mC\big)^{-1} = \mu\Id - \left (\Id- \frac{\mU}{\sqrt{\alpha n}}  \right )  {\mC^{tr}}^{-1}.
    \label{gaussian_map:eq:j_star_asymptotic_high_alpha}
\end{equation}
This expression for the inferred coupling matrix can be inserted in eqn.~(\ref{eq:gc2}) for $\gamma^{cross}$. Carrying out the averages over $\mU$ appearing in $\mJ^*$ and $\mC^{emp}$ we obtain
\begin{equation}
    \gamma^{cross} = \frac{n}{\sum _{i,j}( J_{ij}^*)^2 } \stackrel{\a \rightarrow\infty}{\simeq}  \frac{n}{\sum _{i,j}( J_{ij}^{tr})^2 }.  \label{eq:gamma_cross_infinite_alpha}
\end{equation}
The stronger the interactions in our underlying model, the weaker the regularization that needs to be applied during inference. One way of intuitively understanding this statement is that stronger interactions will \textit{a priori} generate samples (and therefore MAP estimates) with less undesirable variance, and therefore require less smoothing from the regularization.

\subsubsection{$\a \rightarrow 0$ regime}

We now consider the case of very poor sampling. The lowest value of the sampling ratio, $\alpha=\frac 1n$, is reached with a single sample $\vs$ ($p=1$).
The empirical covariance matrix is then easily written as
\begin{equation}\label{eq:MAP2}
    \mC^{emp} = \vs\vs \trs := n \, \vu\vu\trs.
\end{equation}
One eigenvalue of $\mC^{emp}$ is non-zero, and is fixed to $n$ to enforce the spherical constraint\footnote{In numerical experiments on finite size $n$, this constraint is enforced by hand, by rescaling the empirical covariance $\mC^{emp}$ to have a trace exactly equal to $n$. Note that, in the $n\rightarrow \infty$ limit and for $\sigma<1$, this rescaling is not necessary. For $\sigma$ larger than $1$, however, the norm of $\vs$ fluctuates strongly, as $|\vs|^2$ follows a chi-square distribution.}. In other words, the normalized vector $\vu=\vs/\sqrt n$ is the unique non-zero eigenvector of $\mC^{emp}$.

\paragraph{Eigenvalues of $\mJ^*$.} The inferred coupling matrix reads, according to eqns.~(\ref{gaussian_map:eq:MAP_equation}) and (\ref{eq:MAP2}),
\begin{equation}
     \mJ^* = \left[j^*(n)- j^*(0)\right] \vu\vu\trs + j^*(0) \, \Id \ , \label{J_star_small_alpha}
\end{equation}
where the eigenvalues $j^*(c^{emp})$ are given by eqn.~(\ref{gaussian_map:eq:MAP_equation_eigenvalues}). Using $\alpha =\frac 1n$ and expanding in powers of $\frac 1n$, we find
\begin{eqnarray}
j^*(0) &=& -\frac{1}{n\gamma{\mu^*}} + \frac{2}{n^2\gamma^2{\mu^*}^3} + O(n^{-3}). \label{eq:j_star_0_small_alpha}\\
j^*(n) &=&  \frac{1}{2\gamma} \Bigg[ 1 + \gamma{\mu^*} -\sqrt{(\gamma{\mu^*}-1)^2+\frac{4\gamma}{n}}\Bigg] + O(n^{-3}). \label{eq:j_star_n_small_alpha}
\end{eqnarray}
The latter expression can be divided into two cases, depending on whether $\gamma\mu$ is larger or smaller than $1$:
\begin{equation}
j^*(n) =
    \begin{dcases}
    {\mu^*} - \frac{1}{n(1-\gamma{\mu^*})} + \frac{\gamma}{n^2(1-\gamma{\mu^*})^3} + O(n^{-3}) \text{ if $\gamma\mu<1$}\\
     \frac{1}{\gamma} - \frac{1}{n(\gamma{\mu^*}-1)} + \frac{\gamma}{n^2(\gamma{\mu^*}-1)^3} + O(n^{-3}) \text{ if $\gamma\mu>1$.}\\
    \end{dcases}
\end{equation}
which, together with the normalization condition
\begin{equation}
    \frac{n-1}{{\mu^*}-j^*(0)} + \frac{1}{{\mu^*}-j^*(n)} = n,
\end{equation}
yields that:
\begin{equation}
\mu^*(\gamma) =
    \begin{dcases}
    \gamma^{-1/2} \text{ if $\gamma<1$}\\
    1 \text{ if $\gamma{\mu^*}>1$.}\\
    \end{dcases} \label{eq:mu_star_small_alpha}
\end{equation}


\paragraph{Expression for $\gamma^{cross}$.}
We then express the terms appearing in the expression of $\gamma^{cross}$, see eqn.~(\ref{eq:gc2}), in terms of the eigenvalues $j^*(0),j^*(n)$:
\begin{eqnarray}\label{eq:j0jn}
    \sum_{i,j} (J^*_{ij})^2 &=&     j^*(n)^2+ (n-1) \,{j^*(0)}^2\ , \\ \label{eq:j0jn2}
 \sum_{i,j} J^*_{ij}\, C^{emp}_{ij}   &=& n\, \left[j^*(n)- j^*(0)\right] + n \, j^*(0)\ ,
 \\\label{eq:j0jn3}
 \sum_{i,j} J^*_{ij}\, C^{tr}_{ij}   &=& n\, \left[j^*(n)- j^*(0)\right]\, \theta + n \, j^*(0) \ ,
\end{eqnarray}
where we introduced the matrix element
\begin{equation}
    \theta = \frac{1}{n} \sum_{i,j} \vu_i \mC^{tr}_{ij} \vu_j\ .
\end{equation}
Let us consider this quantity in more details. On average over the  sample $\vs(=\sqrt{n}\vu)$, we have:
\begin{equation}
    \left\langle \vu_i \vu_j\right\rangle =\frac{1}{n} \mC^{tr}_{ij},
\end{equation}
and thus
\begin{equation}\label{eq:sum8}
    \left\langle \theta \right\rangle = \frac{1}{n^2} \sum_{i,j} {\mC^{tr}_{ij}}^2 = \frac{1}{n^2} \sum_k {c^{tr}_k}^2.
\end{equation}
Generally, due to the constraint that $\sum_k c^{tr}_k = n$, we find that $\left\langle \theta \right\rangle$ is bounded from below by $1/n$ (when all eigenvalues of $\mC^{tr}$ are equal to $1$), and from above by $1$ (when a single eigenvalue of $\mC^{tr}$ is equal to $n$, and all the other eigenvalues are equal to $0$). It should be noted that, while the result $\langle\theta\rangle>1/n$ is true on average, the value of $\theta$ for an individual sample can be arbitrarily close to $0$. This possibility will be discussed below.

\begin{figure}
    \centering
    \includegraphics[width=\textwidth]{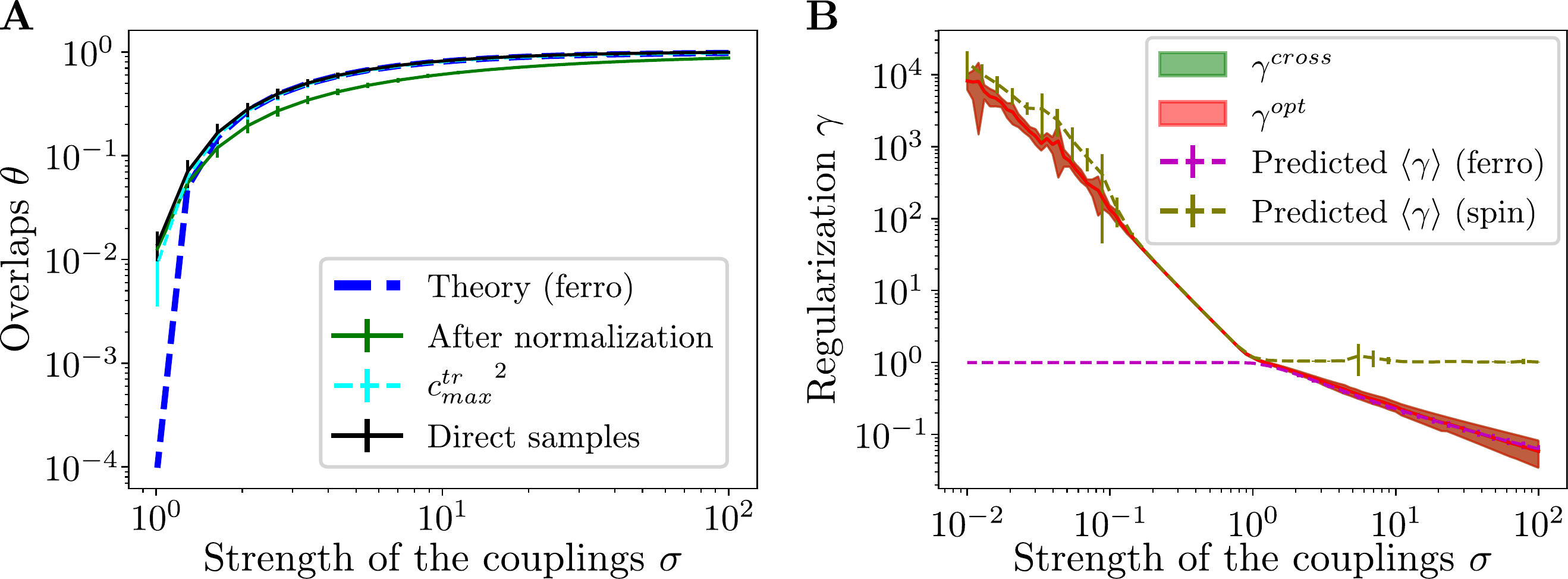}
    \caption[Inference from a single pattern]{Properties of the inference in the low sampling regime $\a=1/n$ and for the random quenched coupling model. \textbf{A}: Value of the overlap $\theta$ as a function of the scale $\sigma$ of the interactions in the ferromagnetic regime $\sigma>1$, see theoretical prediction for $\langle \theta \rangle$ for non rescaled samples in eq.~(\ref{eq:convsig1}). For normalized samples the overlap converges towards $1$ more slowly. \textbf{B}: Comparison between the values of $\gamma^{opt}$ and $\gamma^{cross}$ found numerically   and the predictions in eqns,~(\ref{gaussian_map:eq:small_alpha_spin}) and (\ref{gaussian_map:eq:small_alpha_ferro}), applied to the empirical distribution of "rescaled samples" overlaps. In both panels  error bars represent the variations across 10 choices of the true underlying interaction matrix of the $\theta$, and $\gamma$ is averaged over 100 random draws from the Gaussian model distributioh.}
    \label{gaussian_map:fig:small_alpha}
\end{figure}

Analytical expressions for the average value of $\theta$ can be obtained in the case of random quenched interactions considered in Section~\ref{sec:rqc} by explicitly integrating over the semi-circle eigenvalue distribution, with the results
\begin{equation}
   \langle \theta \rangle = \left\{ \begin{array}{c c c}
 \frac{1}{n(1-\sigma^2)} & \text{if} & \sigma < \sigma_c=1 \ , \\
    \left(1-\frac 1\sigma\right)^2 & \text{if} & \sigma > \sigma_c \ ,
    \end{array} \right. \label{eq:convsig1}
\end{equation}
see Figure~\ref{gaussian_map:fig:small_alpha}A. The last equation comes from the fact that,  when $\sigma$ is larger than $1$, the sum in eqn.~(\ref{eq:sum8}) is dominated by the single macroscopic eigenvalue of $\mC^{tr}$, see eqn.~(\ref{eq:max_eig_ferro}). For the rescaled samples we used in practice, the average value of $\theta$ still goes to $1$ as $\sigma$ increases, but with a gap closer to $\sim \sigma^{-1/2}$.

Let us now summarize the different cases that can be met, see  Figure~\ref{gaussian_map:fig:small_alpha}B:
\begin{itemize}
    \item If $\sigma$ is below $1$, the system is in a disordered phase, and strong regularization is needed. We find that
    \begin{itemize}
        \item if $\theta > 1/n$, the crossing regularization is
                \begin{equation}
                 \gamma^{cross} = \frac{n\theta}{n\theta-1}, \label{gaussian_map:eq:small_alpha_spin}
                \end{equation}
            which is larger than 1. This corresponds to a situation where the sample is slightly informative, and strong regularization is necessary to avoid overfitting.
        \item if $\theta < 1/n$, the two likelihoods never cross, and the optimal regularization appears to be infinite. This corresponds to a situation in which the randomly drawn sample is counter-informative, so that the null answer is better than taking it into account.
    \end{itemize}
    \item If $\sigma$ is above $1$, the system is in the ferromagnetic phase, so that a single sample conveys significant information about the entire  distribution. In that case, we find that for a given (rescaled) sample the crossing regularization is given by
    \begin{equation}
        \gamma^{cross} = (1-\theta)^2,\label{gaussian_map:eq:small_alpha_ferro}
    \end{equation}
    which vanishes when $\sigma\to\infty$.
\end{itemize}

In all cases where $\sigma$ is either very small or very large, the optimal regularization varies strongly from sample to sample.

\subsection{Asymptotic behavior of $\gamma^{opt}$ for $\a \to 0$}

While the $\a \rightarrow \infty$ limit of the optimal regularization is hard to obtain (in particular, because the test likelihood's derivative with respect to $\gamma$ vanishes uniformly), the computation of $\gamma^{cross}$ can be carried out in the low ratio regime, $\a = 1/n$.

We start from the definition of $\gamma^{opt}$:
\begin{equation}
    \frac{\partial L^{test}}{\partial \gamma}(\gamma^{opt}) = 0 = \frac{\partial}{\partial \gamma} \left[\frac 12 \sum_{i,j} J_{ij}^* C_{ij}^{tr} - \log\, Z(\mJ^*) \right] \label{eq:L_test_derivative}
\end{equation}

From eqns.~(\ref{eq:logZ}) and (\ref{J_star_small_alpha}), we have
\begin{equation}
    \log Z(\mJ^*) = \frac{n}{2} \mu^* - \frac{1}{2} \left[ (n-1) \log(\mu^*-j^*(0)) + \log(\mu^*-j^*(n))   \right]
\end{equation}
so that
\begin{equation}
\begin{split}
    \frac{\partial \log\, Z(\mJ^*)}{\partial \gamma} &= \frac{n-1}{2} \frac{\partial_{\gamma} j^*(0)}{\mu^*-j^*(0)} + \frac{1}{2} \frac{\partial_{\gamma} j^*(n)}{\mu^*-j^*(n)}\ .
\end{split}
\end{equation}
In addition, differentiating eqn.~(\ref{eq:j0jn3}) we get
\begin{equation}
\begin{split}
  \frac{\partial }{\partial \gamma}  \sum_{i,j} J_{ij}^* C_{ij}^{tr} &= n\left[\frac{\partial j^*(n)}{\partial \gamma}- \frac{\partial j^*(0)}{\partial \gamma}\right]\theta + n \frac{\partial j^*(0)}{\partial \gamma}
\end{split}
\end{equation}

We now need to evaluate the derivatives $\partial_{\gamma} j^*(0)$ and $\partial_{\gamma} j^*(n)$. From eqns.~(\ref{eq:j_star_0_small_alpha}) and (\ref{eq:mu_star_small_alpha}), at the first order in $n$, we have
\begin{equation}
    \frac{\partial j^*(0)}{\partial \gamma} =  \frac{1}{n \gamma^2 \mu^*} + \frac{\partial_{\gamma}\mu^*}{n \gamma {\mu^*}^2} =\left\{
\begin{array}{c c c}
     \frac{1}{n \gamma^2} &  \text{if}  &\gamma > 1\ ,\\
    \frac{1}{ 2n \gamma^{3/2}} & \text{if} &  \gamma < 1\ .
\end{array}\right.
\end{equation}
Similarly, eqns.~(\ref{eq:j_star_n_small_alpha}) and (\ref{eq:mu_star_small_alpha}) yield
\begin{subnumcases}{ \frac{\partial j^*(n)}{\partial \gamma} =}
    - \frac{1}{\gamma^2} &  if  $\gamma > 1$\ ,\\
    -\frac{1}{2\gamma^{3/2}}  & if  $\gamma < 1$\ .
\end{subnumcases}

We may now conclude our calculation of $\gamma^{opt}$:
\begin{itemize}
    \item If $\gamma > 1$, \begin{equation}
    \frac{\partial L^{test}}{\partial \gamma} = \frac{1}{2\gamma^2} (1-n\theta) + \frac{1}{2 \gamma^2 (\gamma-1)}.
\end{equation}
and therefore this derivative vanishes for
\begin{equation}
    \gamma^{opt} = \frac{n\theta}{n\theta-1},
\end{equation}
which is the same result as found from the $\gamma^{cross}$ computation in equation (\ref{gaussian_map:eq:small_alpha_spin}).
    \item If $\gamma < 1$, \begin{equation}
    \frac{\partial L^{test}}{\partial \gamma}=\frac{n}{4 \gamma^{3/2}} \left[(1-\theta) - \sqrt{\gamma} \right] \ ,
\end{equation}
whose root is given by
\begin{equation}
    \gamma^{opt} =(1-\theta)^2 \ ,
\end{equation}
in full agreement with the result shown in eqn.~(\ref{gaussian_map:eq:small_alpha_ferro}).
\end{itemize}

Therefore, the analytical expressions of $\gamma^{opt}$ and $\gamma^{cross}$ coincide in the undersampled regime (single sample), which provides further support to our conjecture that the values of those two regularizations are equal or very close, as suggested by numerical experiments. Unfortunately, the computation of $\gamma^{opt}$ in the oversampled regime ($\alpha \rightarrow \infty$)  is more complicated, and we were not able to prove that its value converges to the limit found for $\gamma^{cross}$ in eqn.~(\ref{eq:gamma_cross_infinite_alpha}).

\section{Conclusion}

In this work we provided both analytical and numerical evidence for the optimal value of  a $L_2$ penalty term in the likelihood used for Maximum A Posteriori inference of graphical models. In addition to showing that a non-zero optimal regularization always exists, we find a remarkable empirical coincidence between two optimality criteria: the maximization of the test log-likelihood, and the condition that test and generated likelihoods are equal, a natural requirement for a generative model, see Figure~\ref{gaussian_map:fig:noticeable_gammas}. This equality suggests that, while weaker regularizations might give the impression of higher quality generated data (through higher generated likelihoods), stronger regularizations should actually be employed to achieve the best possible model, and the perceived increase in generated likelihood is actually a form of overfitting.

Analytical expressions for the crossing and optimal regularizations could be obtained in the limiting regimes of poor or good sampling. In the latter case, we obtain an explicit expression for the optimal regularization strengths in terms of the average inverse squared couplings between the variables, see eqn.~(\ref{eq:pred}). This prediction remains remarkably accurate over a wide range of parameter value, and even for case of categorical variables (Potts model), while it was established analytically in the case of the Gaussian multivariate model. This result suggest that our study could also be applied to other interesting classes of models, such as Restricted Boltzmann Machines, an extension of Ising/Potts models in which multi-body interactions can be introduced. More generally, it has been known for a long time that Neural Networks benefit from regularization, with extensive research being led on the exact regularization scheme to apply for different tasks (see for example \cite{wan_regularization_2013, zaremba_recurrent_2015,louizos_learning_2018, haarnoja_soft_2018, bartlett_deep_2021}); all approaches exhibit some form of "bias-variance trade off", \textit{i.e.} a phenomenon in which increasing the strength of the regularization reduces the variance of the estimator (\textit{e.g.} by increasing the smoothness of the solutions) but in doing so biases the inference towards a particular subset of solutions; because of this, an optimal value of the regularization exists that balances those two effects, very similarly to what we observed in our simplistic model.

In terms of modeling protein from sequence data our results suggest that the optimal $\gamma$ should neither be proportional to $\frac pn$ nor to $q$, as proposed in previous works  \citep{ekeberg_fast_2014, hopf_mutation_2017}, but is related to the inverse sum of the squared couplings incoming onto residues, see eqn.~(\ref{eq:pred}). In particular, our prediction is that the optimal value for $\gamma$ scales inversely proportional to the number of interacting neighbors on the dependency graph. However, some caution must be brought to this conclusion. The sample size $p$ is not clearly defined for real proteins. The presence of phylogenetic correlations between sequences make the assumption of independent data points only approximate at best. In practice the choices $\gamma=0.01 \frac pn$ \citep{ekeberg_fast_2014} and $\gamma = 0.01\, q$ \citep{hopf_mutation_2017} are qualitatively similar when the number of sequences exceed the protein length by a factor 20, which is not unreasonable for a substantial number of protein families.

Last of all, let us recall that we focused in this work only on Maximum A Posteriori inference, which can be seen as the null temperature limit of Bayesian inference. It is natural to wonder whether our result hold for when sampling the posterior probability at inverse temperature $\beta$:
\begin{equation}
    p_{\beta}(\mJ) \propto e^{ -\beta \left[ \frac{\gamma}{4} Tr(\mJ^2) - \frac{\alpha}{2} Tr(\mJ \mC^{emp}) + \alpha \log Z(\mJ) \right]}\ . \label{gaussian_map:eq:finite_temperature}
\end{equation}
While an in-depth study of the different sampling strategies is out of the scope of this work (see \cite{rubinstein_simulation_2016} for a general overview), we report below numerical and analytical preliminary steps aiming at characterizing this posterior distributions.

\begin{figure}
    \centering
    \includegraphics[width=\textwidth]{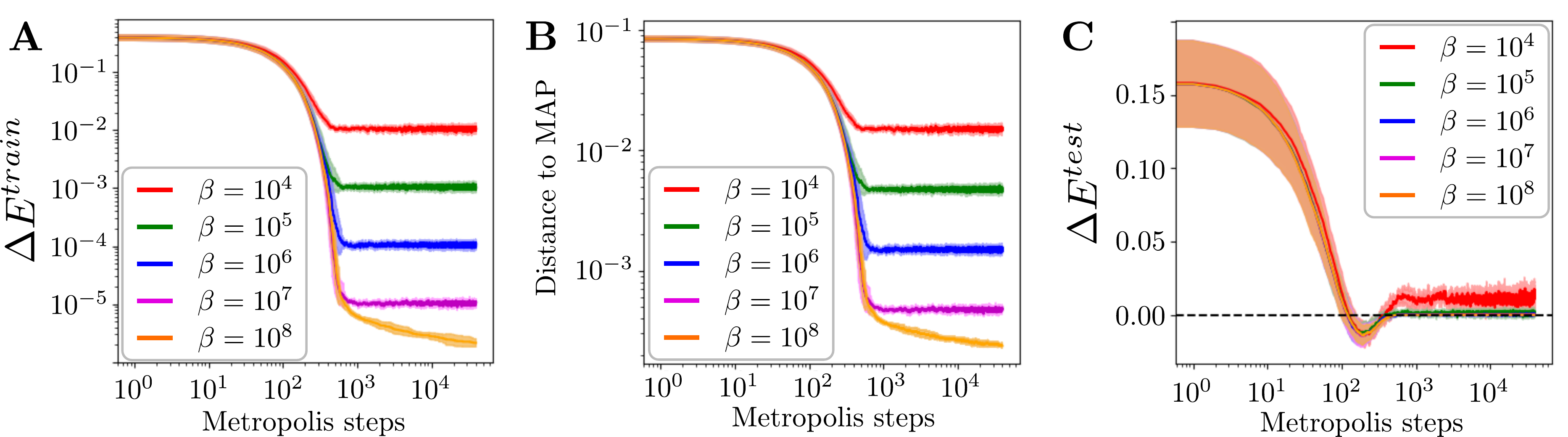}
    \caption[Finite temperature]{Evolution of the train energy, distance to MAP estimator and test energy as a function of the number of Metropolis steps for different values of the temperature. The energies are given relative to the ones of the MAP. For low temperatures and long enough times, the sampled solutions have very close energies to the MAP estimator. At intermediate times, the test energy of the sampled solutions can get lower than the one of the MAP. Higher temperature allow the system to stay in states of higher energy, which are further from the MAP. Figure obtained with $n=20$, $\alpha=5$, $\sigma=0.5$, $\gamma=5$ (larger than the optimal regularization $\gamma^{opt}= 1/\sigma^2=4$).}
    \label{gaussian_map:fig:finite_temperature}
\end{figure}

We performed some preliminary experiments using a simple Metropolis-Hastings algorithm \cite{metropolis_monte_1949} which consists in starting from a random point in the distribution, proposing a small modification and accepting it with probability $p=\min(1, \exp(-\beta \Delta E))$ depending on the associated change in energy, $\Delta E$. In our case, we start from a symmetric Gaussian matrix in which all the entries above the diagonal are independent and have the same mean and variance as the MAP estimator\footnote{This initial choice only affects convergence time, as the Metropolis sampling procedure loses information on the initial conditions after a transient regime.}, and the modifications we propose are the addition of small amplitude, sparse, Gaussian matrices. Since increasing the temperature (hence decreasing $\beta$) can be seen as a way of letting the system explore areas of higher energy, the matrices sampled at higher temperatures will be further away from the MAP solution, which we illustrate in Figure~\ref{gaussian_map:fig:finite_temperature}\textbf{A} and \textbf{B} respectively. While the energy used for sampling is computed using the empirical covariance matrix $\mC^{emp}$, it is also interesting to consider the evolution of a "test" energy, computed using the true covariance matrix $\mC^{tr}$, which will help quantify the generalization property of these solutions. While at long time scales the test energy converges to a value very close to the one of the MAP estimator, there exists an intermediate regime in which the sampled matrices achieve better test energy than the MAP estimator, as seen in Figure~\ref{gaussian_map:fig:finite_temperature}\textbf{C}. Notice, however, that the values of the inverse temperature $\beta$ considered in the simulations are large compared to the canonical inverse temperature, $n$, defined in the posterior probability over $\mJ$, see eqn.~(\ref{eq:Bayes}). The results reported above therefore imply that weak fluctuations of the posterior do not modify the properties of the MAP estimator.

\vskip .5cm
\noindent {\bf Acknowledgements.} We are grateful to J. Tubiana for providing us with his code for Annealed Importance Sampling.

\bibliography{biblio.bib}

\begin{appendices}
\section{Numerical estimation of the  regularization strengths}
\label{app:residuals}

In order to compute the values of $\gamma^{opt}$ and $\gamma^{cross}$ as precisely as possible, we derived two \textit{residuals}, \textit{i.e.} functions of $\gamma$ which are equal to 0 respectively when the test likelihood is optimal, or when the test and generated likelihoods are equal. Similarly to how $\mu^*$ was determined when solving the MAP equation, the roots of those residuals will be minimized using standard convex optimization routines to obtain high precision estimates of the optimal and crossing regularizations.

This approach is easilly illustrated in the case of  the crossing regularization $\gamma^{cross}$. According to eqn.~(\ref{eq:gc2}) the following function $Res^{cross}(\gamma)$ has its root equal to $\gamma^{cross}$:
\begin{equation}
   Res^{cross}(\gamma) := \alpha \frac{\langle \mJ^*(\gamma)\, (\mC^{emp}-\mC^{tr}) \rangle}{\langle \mJ^*(\gamma)^2 \rangle} - \gamma. \label{gaussian_likelihood:eq:gamma_cross_residual}
\end{equation}

For the estimation of the optimal regularization, the computation is more involved and relies on finding the derivative of $L_{test}$ with respect to $\gamma$. Indeed, $\gamma$ is equal to $\gamma^{opt}$ when
\begin{equation}
   Res^{opt}(\gamma) := \frac{\partial L_{test}}{\partial \gamma}
\end{equation}
is equal to $0$.

This derivative can be computed as:
\begin{equation}
\begin{split}
        \frac{\partial L_{test}}{\partial \gamma} &= \frac{1}{2} \sum_{i,j} \frac{\partial J^*_{ij}}{\partial \gamma}C^{tr}_{ij} - \frac{\partial \log Z(\mJ^*)}{\partial \gamma}\\
        &= \frac{1}{2} \sum_{k} \frac{\partial j^*_k}{\partial \gamma}C^{tr,rot}_{k,k} - \frac{\partial \log Z(\mJ^*)}{\partial \gamma},
\end{split}
 \label{gaussian_MAP:eq:first_eq_of_residual_computation}
\end{equation}
where $\mC^{tr,rot}$ is the true correlation matrix after changing the basis to the inference basis in which $\mC^{emp}$ is diagonal.

\noindent We begin by computing
\begin{equation}
\begin{split}
    \partial_{\gamma} j^*_k &=\partial_{\gamma} \left[\frac{1}{2\gamma} \a c_k + \gamma\mu^* - D_k\right]\\
    &= A_k\partial_{\gamma}\mu^* + B_k - \frac{j^*_k}{\gamma},
\end{split}
\end{equation}
where we introduced
\begin{eqnarray}
    D_k &=& \sqrt{(\alpha c^{emp}_k - \gamma \mu^*)^2+4 \alpha \gamma}, \\
    A_k &=& \frac{1}{2}\left(1 - \frac{\gamma \mu^* - \alpha c^{emp}_k}{D_k}\right), \\
    B_k &=& \frac{1}{\gamma}\left(\mu^* A_k -\frac{\alpha}{D_k}\right).
\end{eqnarray}

\noindent Then, we have that
\begin{eqnarray}
   \partial_{\gamma} \log Z &=& n\partial_{\gamma} \mu^* - \frac{1}{2} \sum_k \frac{\partial_{\gamma} \mu^*  - \partial_{\gamma} j^*_k}{\mu^* - j^*_k}.
\end{eqnarray}

\noindent Finally, we can compute $\partial_{\gamma}\mu$ by first noting that:
\begin{equation}
    \frac{1}{2} \sum_{k} \frac{1}{\mu^*- j^*_{k}} = 1,
\end{equation}
hence
\begin{equation}
    \sum_{k} \frac{\partial_{\gamma} \mu^* - \partial_{\gamma} j^*_{k}}{(\mu^*- j^*_{k})^2} = 0,
\end{equation}
and therefore
\begin{eqnarray}
    \partial_{\gamma} \mu &=&
      \left[ \sum_k \frac{\partial_{\gamma} j^*_{k}}{(\mu^*- j^*_{k})^2} \right] / \left[\sum_k\frac{1}{(\mu^*- j^*_{k})^2} \right]\\
     \partial_{\gamma} \mu^* &=& \left[ \sum_k \frac{A_k\partial_{\gamma}\mu^* + B_k - j^*_k/\gamma}{(\mu^*- j^*_{k})^2} \right] / \left[\sum_k\frac{1}{(\mu^*- j^*_{k})^2} \right] \\
      \partial_{\gamma} \mu^* \left[\sum_k\frac{1-A_k}{(\mu^*- j^*_{k})^2} \right] &=& \left[ \sum_k \frac{B_k - j^*_k \gamma}{(\mu^*- j^*_{k})^2} \right]
\end{eqnarray}

\noindent which finally yields:
\begin{equation}
    \partial_{\gamma}\mu^* = \left[\sum_k  \frac{B_k - j^*_k/\gamma}{(\mu^*-j^*_k)^2} \right] / \left[\sum_k  \frac{1 - A_k}{(\mu^*-j^*_k)^2} \right].\label{gaussian_MAP:eq:last_eq_of_residual_computation}
\end{equation}

Putting together eqns.~(\ref{gaussian_MAP:eq:first_eq_of_residual_computation}) to~(\ref{gaussian_MAP:eq:last_eq_of_residual_computation}) yields an explicit expression for the derivative of $L_{test}$ with respect to $\gamma$, which is exactly the residual $Res^{opt}(\gamma)$ whose root gives the value of $\gamma^{opt}$.

\end{appendices}

\end{document}